%% file: main.tex
\documentclass[letterpaper, 10 pt, conference]{ieeeconf}  
\usepackage{times}  
\usepackage{helvet} 
\usepackage{courier}  
\usepackage[hyphens]{url}  
\usepackage{graphicx} 
\usepackage{graphicx}  
\usepackage{graphicx}
\usepackage{amsmath}

\usepackage{hyperref}
\usepackage{url}
\usepackage{amsmath,amsfonts,amssymb,mathrsfs}
\usepackage{psfrag,graphicx,epsfig,epsf}
\usepackage[ruled,linesnumbered,vlined]{algorithm2e}
\usepackage{fancyhdr}
\usepackage{wrapfig}
\usepackage{subfigure}
\usepackage{mathtools}
\usepackage{url}
\usepackage{color}
\usepackage{verbatim}
\usepackage{xspace}
\usepackage{float}

\usepackage{algpseudocode}
\usepackage{cite}

\IEEEoverridecommandlockouts                              

\title{\LARGE \bf
Learning Category-Level Manipulation Tasks from Point Clouds\\ with Dynamic Graph CNNs}

\author{
Junchi Liang  and Abdeslam Boularias$^{1}$
\thanks{$^{1}$The authors are with the Department of Computer Science of Rutgers University, Piscataway, New Jersey 08854, USA.
        {\tt\footnotesize \{jl2068, ab1544\}@cs.rutgers.edu}}%
}

\begin{document}

\maketitle
\thispagestyle{empty}
\pagestyle{empty}

\begin{abstract}

This paper presents a new technique for learning category-level manipulation from raw RGB-D videos of task demonstrations, with no manual labels or annotations. Category-level learning aims to acquire skills that can be generalized to new objects, with geometries and textures that are different from the ones of the objects used in the demonstrations. 
We address this problem by first viewing both grasping and manipulation as special cases of {\it tool use}, where a tool object is moved to a sequence of key-poses defined in a frame of reference of a target object. 
Tool and target objects, along with their key-poses, are predicted using a dynamic graph convolutional neural network
that takes as input an automatically segmented depth and color image of the entire scene.
Empirical results on object manipulation tasks with a real robotic arm show that the proposed network can efficiently learn from real visual demonstrations to perform the tasks on novel objects within the same category, and outperforms alternative approaches. 
\end{abstract}

\input{introduction}

\input{related}

\input{problem}
\input{proposed}

\input{experiments}
\input{conclusion}

\bibliographystyle{IEEEtran}
\bibliography{bib,iros20}
\end{document}

%% file: introduction.tex
\section{Introduction}

Current robotic manipulation systems that are deployed in real-world environments rely on precise 3D CAD models of the manipulated objects. These models are used by planning algorithms in simulation to carefully select actions of the robot to apply on the objects in order to accomplish a pre-defined task, such as pick-and-place, assembly, transferring liquid content, and surface painting. 3D CAD models are obtained by manually placing standalone objects on rotary tables and scanning them.
Moreover, these objects need to be detected from images, typically through the use of convolutional neural networks that are trained by manually selecting and labeling the objects in training images. This tedious human labor severely limits the applicability of manipulation system outside of controlled environments, where robots are faced with a large variety of object types, shapes and textures, which makes it virtually impossible for robots to always rely on prior models. Moreover, despite recent progress in model-based planning algorithms, search for manipulation actions in a physics-based simulation that involves contacts and collisions is still computationally expensive. 

An increasingly popular solution to the problems mentioned above is to learn {\it category-level} manipulation skills. A sequence of images showing how to perform a certain manipulation task is provided to the robot by a human demonstrator, or collected by the robot itself through trial-and-error. The robot is then tasked with learning a manipulation policy that can be tested on objects other than those used in the demonstrations. There are typically two types of methods for solving this problem: {{\it end-to-end} techniques that train a neural net to map pixels directly to torques, and {\it modular} techniques that decompose the problem into first reasoning over 6D poses of objects then solving the inverse kinematics and dynamics problem in order to control the robotic arm.  While end-to-end techniques are appealing due to their simplicity and flexibility, they still require large sets of training data and suffer from limited ability to generalize to novel objects. Therefore, the method proposed in this work subscribes to the modular type of robot learning techniques.  

\begin{figure}[t]
    \begin{center}
     \begin{subfigure}{}
     \includegraphics[width=0.49\textwidth]{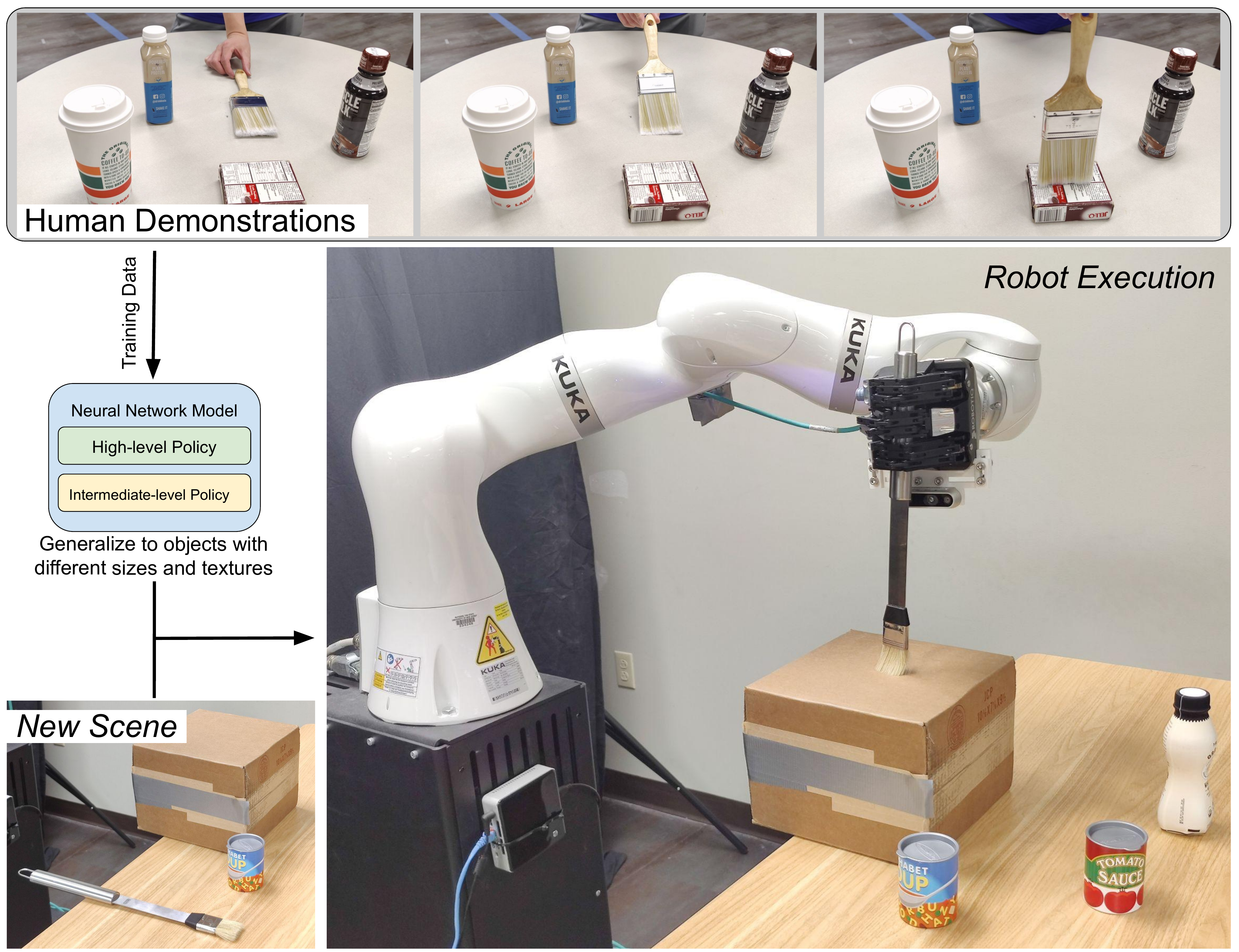}
     \end{subfigure}
    \end{center}
     \caption{\small System overview and robotic setup used in the experiments. In this example, the system is trained by unlabeled visual demonstrations to pick up a paint-brush in a specific manner, and to place it in a specific configuration relative to another object. The robot is tasked with repeating the demonstrated behaviour on a new scene containing novel objects with different sizes and texture.}
\label{fig:system_overview}
\end{figure}

Prior works in modular robot learning were mostly focused on grasping problems~\cite{DBLP:conf/iros/BoulariasKP11,Detry2013,Lenz2013,7139793,Yan-2018-113286,DBLP:conf/iccv/MousavianEF19}.  While there are numerous learning-based techniques that generalize grasping skills to novel objects, only a few recent ones have been shown capable of generalizing other manipulation skills, such as placing, to novel objects~\cite{5509439,seita2019deep,Manuelli2019kPAMKA,DBLP:conf/corl/FlorenceMT18,9363610,e4e03a5d1bf545c29f8c26b824a57068, DBLP:journals/corr/abs-1910-11977, vecerik2020s3k}. However, most of these category-level techniques rely on annotated images wherein a human expert manually specifies {\it keypoints} on training objects. A neural network is then used to predict these task-specific keypoints on new images of unknown objects within the same category. Some techniques, such as~\cite{DBLP:journals/corr/abs-1910-11977} overcome this requirement by using a self-supervised robotic system to collect and annotate data, but this still requires long hours of real robotic manipulation which can be unsafe or less efficient if no guided exploration is provided. 

In this work, we propose a new technique for learning category-level manipulation skills from unlabeled RGB-D videos of demonstrations. The proposed system is fully autonomous, and does not require any human feedback during training or testing besides the raw demonstration videos. The setup of the system includes a support surface wherein a number of unknown objects are placed, which includes task-irrelevant distractive objects. The system is composed of a high-level policy that selects a tool and a target from the set of objects in the scene at each step of the manipulation, an intermediate-level policy that predicts desired 6D keyposes for the robot's wrist, and a low-level policy that moves the wrist to the keyposes. The system employs a dynamic graph convolutional neural network that receives as inputs partial point clouds of the objects. A local frame of reference is computed  for each object based on the principal component analysis of its point cloud. The 6D keyposes predicted by the network are in the frame of the predicted target object. 

The proposed system is tested on a real robot with real demonstrations of four different tasks: two variants of stacking, simulated pouring, and simulated painting. Simulated pouring and painting are performed with real objects but with no liquid yet, for safety of the robot. The same architecture and parameters are used for the four different tasks. 

The key novel contributions of this work are as follows. (1) An efficient {\it new  architecture} for learning category-level manipulation tasks. A key feature of this architecture is the capability to generate trajectories of the robot's end-effector according to the 3D shapes of the objects that are present in the scene. The proposed architecture can thus generalize to objects with significantly different sizes and shapes. This is achieved through the use of a dynamic version of graph neural networks, wherein the graph topology is learned based on the demonstrated task. Moreover, the proposed architecture can be used in scenes that contain an arbitrary number of objects. This is in contrast with most existing manipulation learning methods, which are limited to objects of similar dimensions. 
(2) An {\it empirical study} comparing various architectures for learning manipulation tasks, using real objects and robot. The study shows that, amongst the compared methods, a dynamic GNN provides the most data-efficient architecture for learning such tasks. Furthermore, the same hyper-parameter values can be used to learn different tasks.
(3) A {\it new formulation} of the problem by representing states as lists of relative 6D poses of object pairs. The proposed formulation also {\it unifies} the problems of grasping and manipulation by treating the end-effector as one of the objects in the scene, and viewing grasping as a special type of tool-use.
(4) A {\it fully self-supervised} learning pipeline that does not require any manual annotation or manual task decomposition, in contrast with existing category-level techniques~\cite{DBLP:journals/ral/SchmidtNF17,5509439,seita2019deep, Manuelli2019kPAMKA,DBLP:conf/corl/FlorenceMT18,9363610,e4e03a5d1bf545c29f8c26b824a57068,DBLP:journals/corr/abs-1910-11977,vecerik2020s3k}.  



%% file: related.tex
\section{Related Work}
\label{sec:related}	


\noindent {\bf Category-Level Manipulation.} Recently, several techniques have been devised for learning to generalize robot manipulation skills to new objects in the same category, such as cups of different sizes, shapes and texture. The new objects do not have  CAD models, and no new data or additional robot-object interactions are needed during testing. Most of these techniques use semantic 3D keypoints to represent objects~\cite{DBLP:journals/ral/SchmidtNF17}. The main advantage of this representation is the ability to specify any manipulation task as a function of specific keypoints on the manipulated objects, which can easily be predicted on new objects, in contrast with more classical methods that rely on CAD models and specify tasks in terms of rigid 6D poses. 
Keypoints were initially proposed as grasp points for the manipulation of deformable objects~\cite{5509439,seita2019deep}. The technique presented in~\cite{Manuelli2019kPAMKA} consists in manually annotating task-relevant keypoints on a large number of training objects, and trains an {\it integral neural network} to align the keypoints between intra-class object instances. Manipulation skills, expressed as constraints over keypoint positions, can then be transferred to new objects. The need for tedious human annotation was removed in~\cite{DBLP:conf/corl/FlorenceMT18} through the use of a self-supervised camera-mounted robotic system that automatically re-arranges objects while attempting to grasp them. Contrastive learning was then employed to learn feature descriptors of 2D image points, which were then shown to be efficient for learning grasps~\cite{9363610}. Similar object-centric dense descriptors were used for learning pick-and-place tasks from demonstrations~\cite{e4e03a5d1bf545c29f8c26b824a57068}. In~\cite{DBLP:journals/corr/abs-1910-11977}, task-specific keypoints are learned from self-supervised robot interactions instead of human annotations. Semantic category-level 3D keypoints were also trained in~\cite{vecerik2020s3k} using a combination of self-supervised training and human annotations.

\noindent {\bf Category-Level Grasping.} Grasping techniques are typically categorized in two main categories: {\it analytical} and {\it data-driven}~\cite{10.1109/TRO.2013.2289018}. 
Analytical approaches require accurate CAD and mechanical models of objects to simulate {\it force-closure} or {\it form-closure} grasps~\cite{grasping,liang2019pointnetgpd,doi:10.1177/0278364912442972}. A CAD model of a novel object is usually obtained by using a rotary table or multiple cameras that cover all angles, it is difficult to obtain CAD models on the fly from a single 2.5 depth image.
Moreover, mass and friction coefficients are generally difficult to infer from vision alone. 
Consequently, most recent grasping techniques focus on learning directly grasp success probabilities from data~\cite{DBLP:conf/iros/BoulariasKP11,Detry2013,Lenz2013,7139793,Yan-2018-113286,DBLP:conf/iccv/MousavianEF19}. More recent learning techniques were adapted to grasp objects in clutter~\cite{DBLP:conf/aaai/BoulariasBS14,Pas2015UsingGT,DBLP:conf/icra/PintoG16,pmlr-v78-mahler17a,mahler2017dexnet,kalashnikov2018qtopt}. Recent proposed methods include training a hierarchy of supervisors from demonstrations to grasp objects in clutter~\cite{7743488}, and convolutional neural networks, such as Dex-net 4.0~\cite{mahler2019learning}, that are trained to detect grasp 6D poses in point clouds~\cite{DBLP:journals/corr/PasGSP17}.  
These data-driven techniques, however, provide grasps that are valid only for picking up objects, without taking into account the desired manipulation task. 
In our proposed method, the robot learns from demonstration different types of grasps for different manipulation tasks, such as pouring or stacking. Goal-conditioned grasping was considered in recent works, such as~\cite{DBLP:journals/corr/abs-2109-09163}. A dataset of 3D CAD models in required in~\cite{DBLP:journals/corr/abs-2109-09163} to learn grasps in simulation. Our method requires instead only a small number of demonstration videos without the need for CAD models.
Moreover, our method takes as inputs the 2.5D point clouds of the hand and of the object, so that it can generalize to new hands and new objects within the same category. Finally, the proposed method unifies the grasping and manipulation problems and uses the same architecture to predict 6D keyposes for grasping, stacking, pouring, and other manipulation tasks.

%% file: problem.tex
\section{Problem formulation}
\label{sec:problem}
A human demonstrates a grasping and manipulation task that involves picking up a {\bf tool} object from a tabletop that contains an arbitrary set of objects of various types, and moving the object to a position and rotation with respect to a {\bf target} object. Examples of such tasks include: stacking, where the demonstrator picks up a box and puts it on top of a second box; pouring, where the demonstrator picks up a cup or a bottle and places it in a position and rotation relative to another container so that liquid can flow between the two objects. No priors or 3D models of the objects are given, the objects are completely unknown. From only raw RGB-D videos of the demonstrations, and without any labeling or annotation, a robot is tasked with re-producing the tasks on new scenes containing only new objects and not including any of the objects that appeared in the demonstrations. We make one mild assumption: the support surface is flat and the objects are physically separated from each other at the beginning of each task. Therefore, the scene can be segmented automatically by detecting the support surface with RANSAC and removing it from the point cloud. 

We denote the set of demonstrations by $T = \{\tau^1, \dots, \tau^n \}$, wherein each demonstration $\tau^i = (s^i_0, s^i_1, \dots s^i_h)$ is a sequence (or trajectory) of recorded scene states $s^i_t$ at different time-steps $t\in\{0,\dots,h\}$. Note that different demonstration trajectories can have different lengths. A state is a tuple $s^i_t = (<P^i_{0,t},I^i_{0,t}>, <P^i_{1,t},I^i_{1,t}>, \dots, <P^i_{m,t},I^i_{0,t}>)$, wherein each element $P^i_{j,t}$ is a point cloud that corresponds to an object $j$ at time $t$, and $I^i_{j,t}$ is its cropped RGB image. The list of objects in the scene includes the hand of the human demonstrator during training, and the robotic hand during testing. These two are treated similarly to the other objects, as we argue in this work that grasping is only a special case of tool use, with the hand being the tool object. In the first time-step $t=0$ of a demonstration $\tau^i$, point clouds $\{P^i_{j,0}\}_{j=0}^{m}$ are obtained by segmenting the scene's depth image. Each point cloud is then tracked over time, and updated based on the images received at time-steps $t\in\{t,\dots,h\}$. 
The sets of objects used in the demonstrations and during testing are denoted by $\mathcal O_{\textrm{training}}$ and $\mathcal O_{\textrm{testing}}$, respectively, with $\mathcal O_{\textrm{training}} \cap \mathcal O_{\textrm{testing}} = \emptyset$. At the beginning of each demonstration, objects are randomly drawn from $\mathcal O_{\textrm{training}}$ and placed on the support surface. Therefore, each scene contains a number of ``distraction'' objects that are not relevant to the demonstrated tasks. Presence of distraction objects reflects how real-world manipulation tasks are performed in uncontrolled environments. 

In a scene $s_t = (<P_{0,t},I_{0,t}>, \dots, <P_{m,t},I_{0,t}>)$, each point cloud $P_{i,t}$ is assigned an intrinsic frame of reference $X_{i,t} = (c_{i,t},\Vec{x}_{i,t},\Vec{y}_{i,t},\Vec{z}_{i,t})$, wherein $c_{i,t}$ is the 3D centroid of the point cloud, and $\Vec{x}_{i,t},\Vec{y}_{i,t}$ and $\Vec{z}_{i,t}$ are the principal, secondary and tertiary axes of the 3D point cloud of object $j$ at time $t$, all expressed in the camera's coordinates system. The intrinsic frames of reference are computed automatically by performing a PCA on $P_{i,t}$.

A {\bf high-level policy}, denoted by $\pi_{h}$, is used to select a pair of (tool, target) objects from the set of objects present in the scene. This policy receives as input state $s_t = (P_{0,t}, \dots, P_{m,t})$ and returns $o_{\textrm{tool}} \in \{0,\dots,m\}$ and 
$o_{\textrm{target}} \in \{0,\dots,m\}$. A {\bf intermediate-level policy}, denoted by $\pi_{m}$, is used to select a desired {\bf keypose} $K \in \mathbb{R}^3\times \mathbb{SO}(3)$ 
of the robotic arm's wrist. The intermediate-level policy takes as input state $s_t$ as well as the (tool, target) objects returned by the high-level policy. Finally, we denote by $\pi_{l}$ a {\bf low-level policy}. The policy receives as inputs the current robot configuration in a world coordinates frame, denoted by $c_t \in \big(\mathbb{R}^3\times \mathbb{SO}(3)\big)^J$, where $J$ is the number of joints of the robot arm/hand, in addition to a desired keypose $K$ 
of the robotic arm's wrist. The policy returns changes $\Delta c_t$ to apply on $c_t$ to move the robot's wrist to keypose $K$.


%% file: proposed.tex
\section{Proposed Approach}
\subsection{Overview}
We start by providing an overview of the proposed system. 
The system is fully autonomous. The same architecture and parameter values are used for all the different tasks performed on unknown novel objects. The only difference between the different manipulation tasks, such as pouring and stacking, are the video demonstrations provided to train the system for each type of manipulation individually. 

The system receives as input at each step an RGB-D image of the scene containing an end-effector and a number of unknown objects. The system returns a pair of tool and target objects, identified by their point clouds in the coordinates system of the camera and by their local frames computed by the system.  Once the tool and target objects are selected by the system, it returns a 6D key-pose that indicates where the tool's frame of reference should be placed relative to the target's frame. The system also returns a sequence of low-level actions that move the robot's joints so that the tool is placed in the returned key-pose. 

At the beginning of a task, the system always selects the robotic hand as the tool object. The target object selected by the system is one of the many unknown objects present in the scene, which contains also distraction objects to test the system's ability to recognize task-related objects. In a pouring task for example, the first target selected by the system is a bottle that the robot needs to grasp. In a small-on-large stacking task, the first target would be the second-largest box. We assume that the robotic gripper is initially open, and closes once it reaches the 6D keypose generated by the system. Once an object is grasped, which is detected by the system from the point cloud input, the system returns a different pair of (tool, target) objects. In our pouring example, the grasped bottle becomes now the new tool object, and a second object (a cup, for example) is returned by the system as the new target object. In the small-on-large stacking example, the grasped box becomes the tool and the largest box becomes the target object. This process is repeated until the task is performed.

\subsection{Dynamic Graph CNNs}
\label{sec:dgcnn}
At the heart of the proposed system lies a Dynamic Graph Convolutional Neural Network (DGCNN)~\cite{abs-1801-07829}. DGCNN is trained to extract task-relevant shape features of an object. An object is given as a partial point cloud $P = \{p_0,\dots, p_l\}$ wherein $p_i\in\mathbb{R}^3$ are the coordinates of a point in the camera frame. Each point is connected to its $k$-nearest neighbors in $P$. The points in $P$ and their connections form a graph structure, which is given as input to the DGCNN. The first layer of the network performs an edge convolution on the object and returns a set of feature vectors $X = \{x_0,\dots, x_l\}$, one for each point  $p_i\in P$. 
A feature vector $x_i\in X$ is computed as $x_i = \max_{p_j \in k\textrm{NN}(p_i)} h_{\Theta}(p_i,p_j)$ where $h_\Theta$ is a function parameterized by $\Theta = (\theta,\phi)$ and defined as 
$h_{\Theta}(p_i,p_j) = \textrm{ReLU} \big( \theta . (p_i - p_j) - \phi . p_i  \big)$. 
Set $X$ of feature vectors returned from the first layer is used to form a new graph that is given to the second layer, wherein the same operations are repeated using different values for weights $\Theta$. And so on, this process is repeated for a number of layers, followed by fully connected layers that return a feature descriptor of the entire object. 
In each layer of the network, the structure of the graph is dynamically re-defined by connecting each point $x\in X$ to its $k$-nearest neighbors from $X$. This is different from standard  graph CNNs, where the graph structure is defined in the input and kept fixed inside the network. The resulting architecture is more expressive than regular GCNNs, as points from distant parts of an object can become neighbors in later layers, depending on the task.



\subsection{Architecture}

\begin{figure}
    \begin{center}
     \begin{subfigure}{}
     \includegraphics[width=0.49\textwidth]{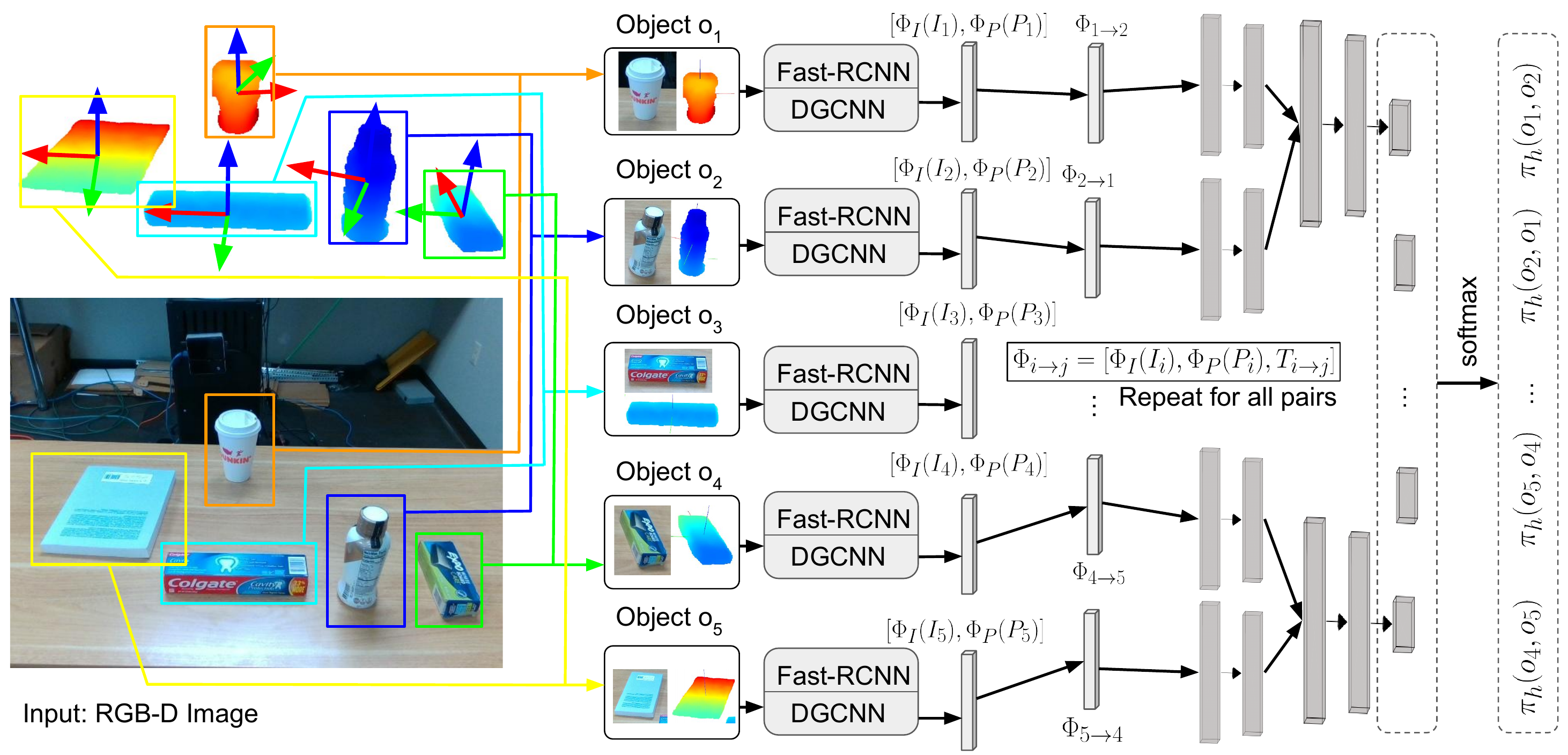}
     \end{subfigure}
    \end{center}
     \caption{High-level policy}
\label{fig:high_level_policy}
\end{figure}

{\bf High-level policy.} The high-level policy receives as input a candidate pair of objects $(i,j)$ and returns $\pi_h(i,j)$, the probability that $(i,j)$ is indeed the pair of tool and target objects needed for performing the manipulation task that the system was trained on. This operation is repeated on all pairs of objects in the scene, and the pair that receives the highest probability $\pi_h(i,j)$ is selected and forwarded to the intermediate policy.  

To compute $\pi_h$, we start by extracting features of each pair $(i,j)$ of the objects present in the scene. For object $j$ paired with object $i$, we extract a descriptor vector $\Phi_{j \rightarrow i}$ of size $1548$ from its point cloud $P_j$ and corresponding cropped RGB image $I_j$. The descriptor vector is defined as $\Phi_{j \rightarrow i}=[ \Phi_{I} (I_j), \Phi_{P} (P_j) , T_{j \rightarrow i}]$. The first component $\Phi_{I} (I_j)$ is a vector of size $1024$ obtained from Fast-RCNN~\cite{Girshick_2015_ICCV} for the RGB features. The second component $\Phi_{P} (P_j)$ is a vector of size $512$ returned by the DGCNN module, as explained in Sec.~\ref{sec:dgcnn}, for the shape features. The last component $T_{j \rightarrow i}$ is a vector of size $12$ that represents the transformation (translation and rotation) of $X_j$ to $X_i$, wherein $X_j$ and $X_i$ are respectively the intrinsic frames of reference of objects $j$ and $i$, computed after performing a PCA on each of their point clouds (Sec.~\ref{sec:problem}). 
Instead of keeping the camera's coordinates system, we take advantage of our pairing input structure, and express an object's PCA pose in the coordinates system of its counterpart, object $i$. With this input, the network can avoid irrelevant influences from different camera poses and focus on motions between the target and the tool objects.

The backbone of the high-level policy is a class-agnostic network that takes as inputs $\Phi_{j \rightarrow i}$ and $\Phi_{i \rightarrow j}$ and returns a score $F_{\theta}(\Phi_{j \rightarrow i}, \Phi_{i \rightarrow j}) \in \mathbb{R}$ for every candidate pair $(i,j)$, wherein $\theta$ are the network's parameters. From these outputs, we define $\pi_h(i,j) = \textrm{softmax}_{(i',j')}F_{\theta}(\Phi_{j' \rightarrow i'}, \Phi_{i' \rightarrow j'})$ by normalizing over all pairs in the scene, and we set $(o_{\textrm{tool}},o_{\textrm{target}}) = \arg\max_{(i,j)} \pi_h(i,j)$.

Because the policy's probabilities $\pi_h(i,j)$ are obtained by normalizing the network's outputs over all pairs in the scene, the size of the network is decoupled from the number of objects and their order. In contrast, an ordinary classifier requires a fixed number and ordering of output classes. So our model is more scalable and compact. The backbone $F_{\theta}$ is composed of encoders consisting of fully connected layers for tool candidate descriptor, $\Phi_{j \rightarrow i}$, and target candidate descriptor, $\Phi_{i \rightarrow j}$, separately. The two branches return two encoding vectors, one for the tool and another for the target. The two encoding vectors are concatenated and provided as input to hidden fully connected layers, which finally output a scalar. In this design, $F_{\theta}$ receives only features for target and tool candidates and has no class-specific architecture or parameters, so it does not require predefined categories and it can generalize to objects with similar shape and appearance without manual labeling. It can also handle scenes with an arbitrary number of objects.

\begin{figure}
    \begin{center}
     \begin{subfigure}{}
     \includegraphics[width=0.49\textwidth]{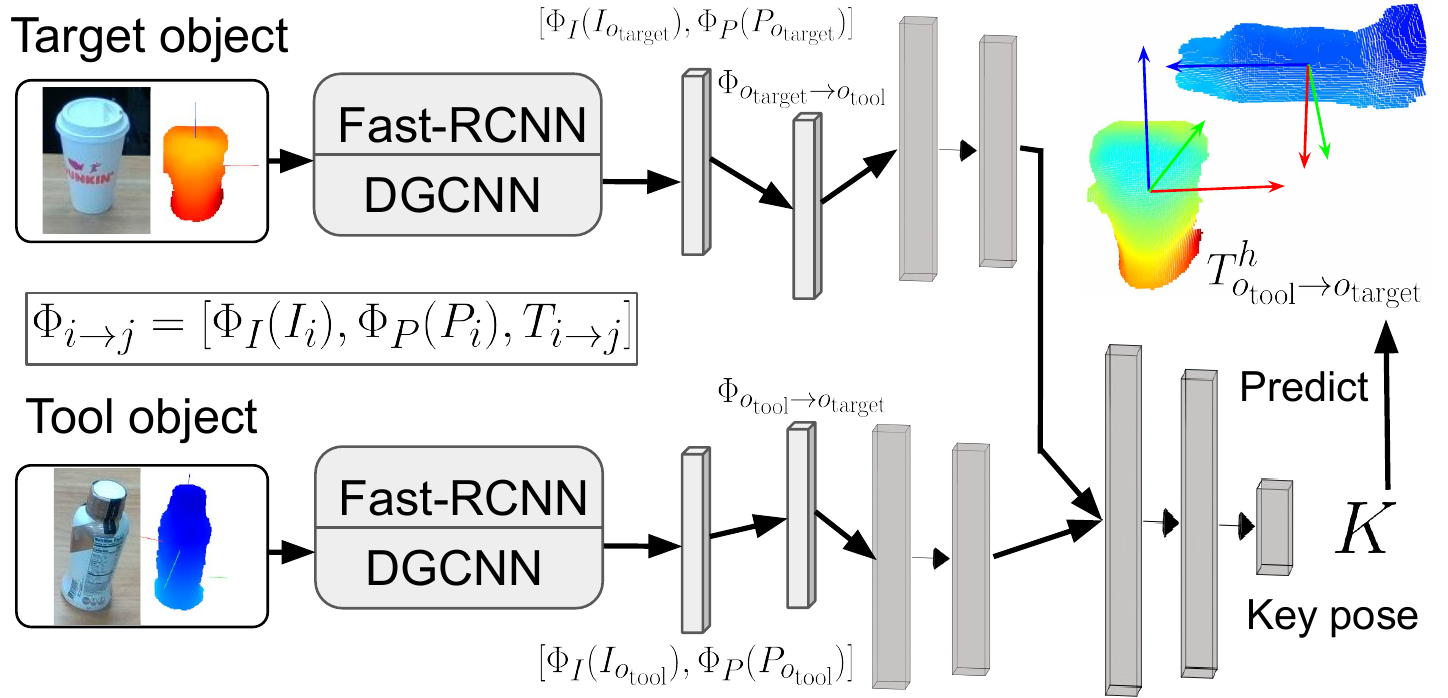}
     \end{subfigure}
    \end{center}
     \caption{Intermediate-level policy}
\label{fig:intermediate_level_policy}
\end{figure}

\begin{figure}
    \begin{center}
     \begin{subfigure}{}
     \includegraphics[width=0.4\textwidth]{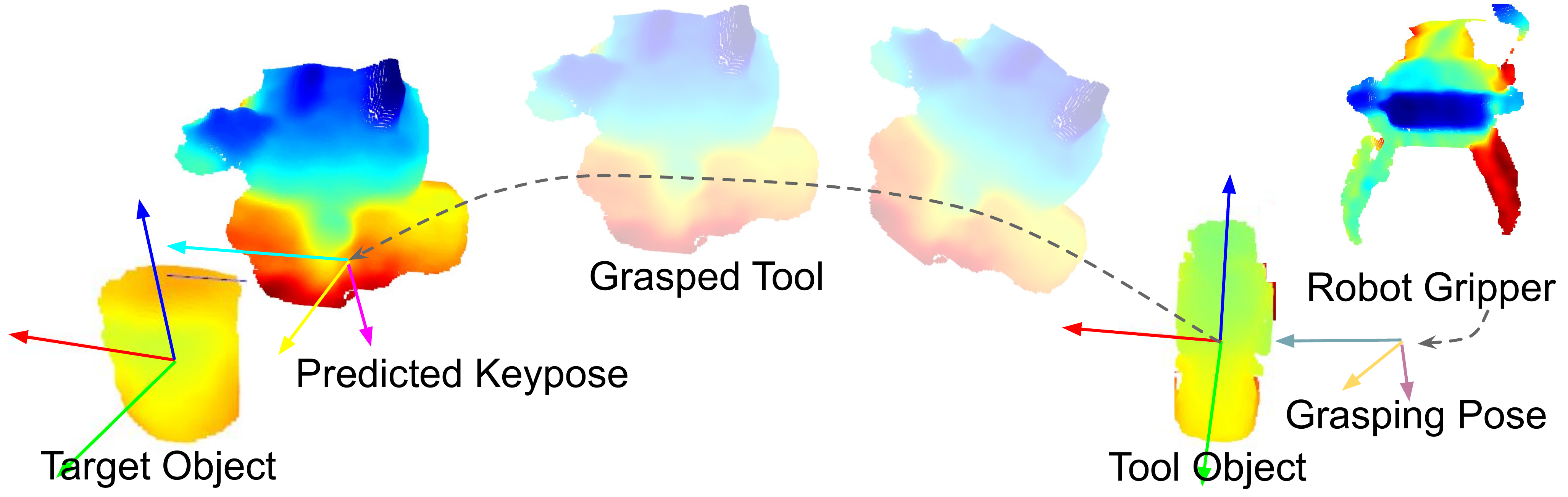}
     \end{subfigure}
    \end{center}
     \caption{Low-level policy}
\label{fig:low_level_policy}
\end{figure}

\begin{figure*}[t!]
    \begin{center}
     \begin{subfigure}{}
     \includegraphics[width=0.49\textwidth]{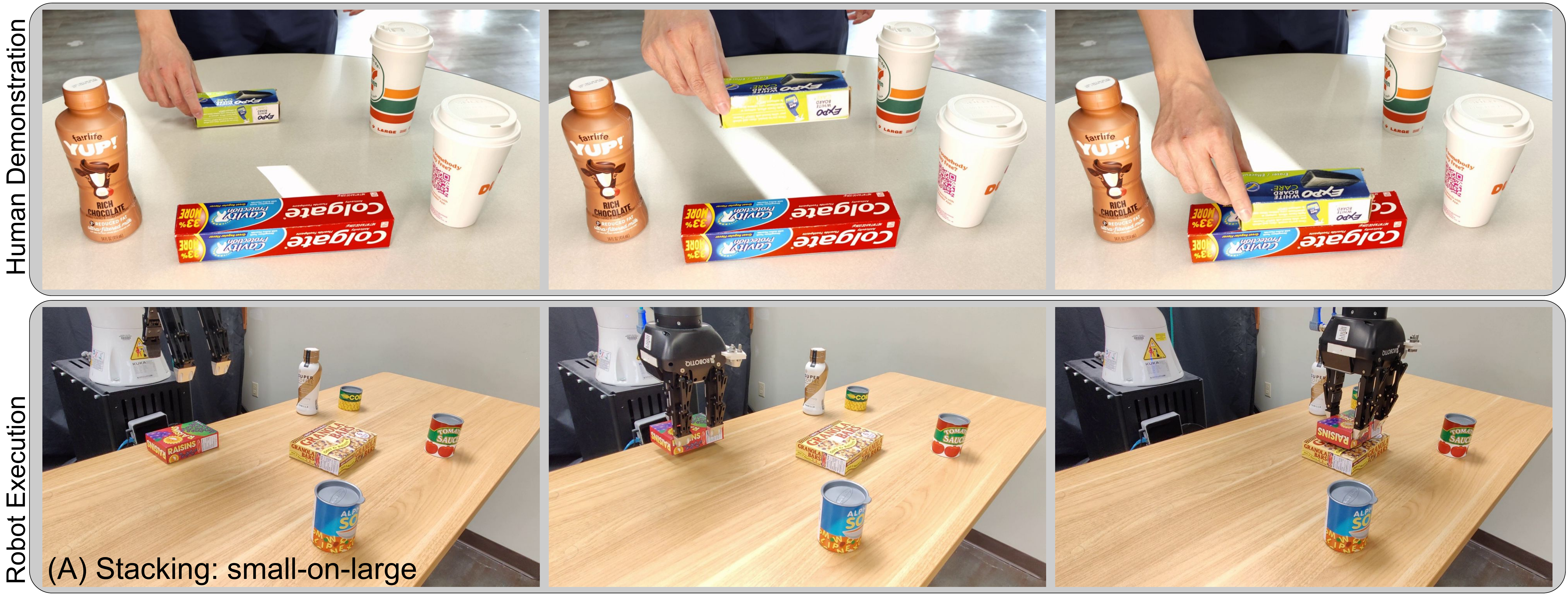}
     \end{subfigure}
     \hspace{-4mm}
    \begin{subfigure}{}
     \includegraphics[width=0.49\textwidth]{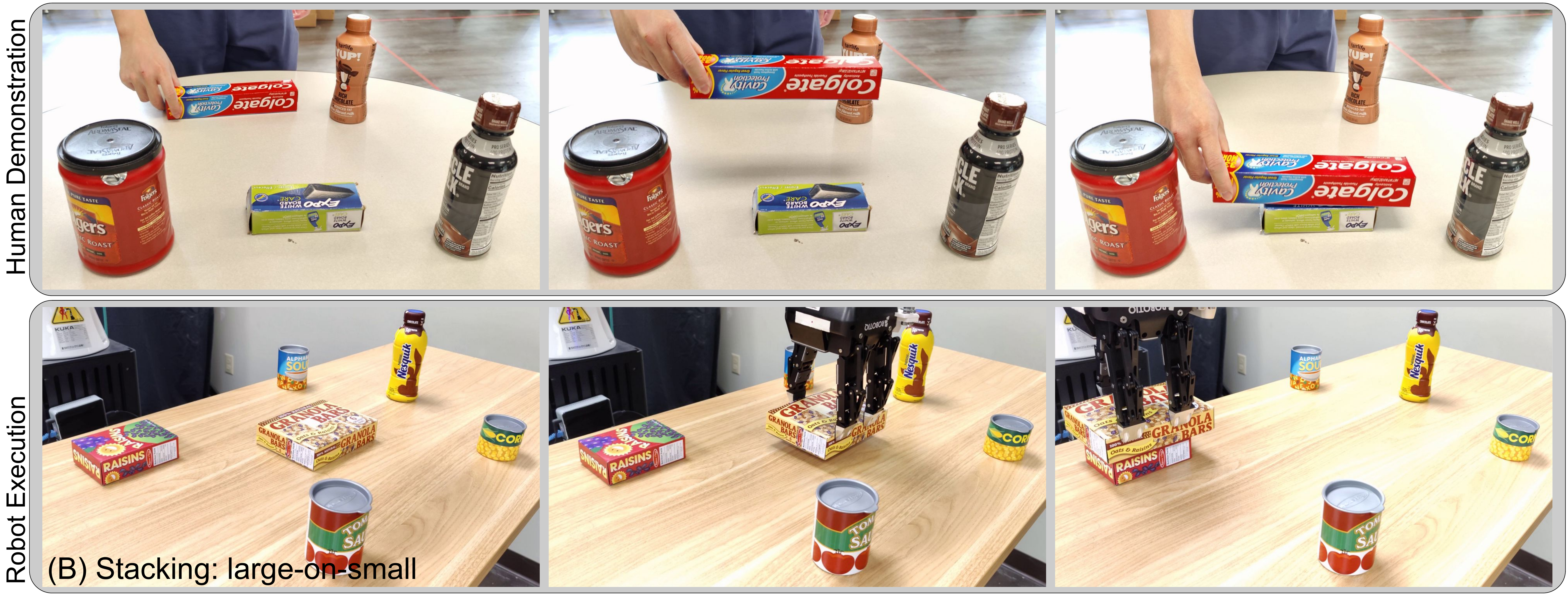}
     \end{subfigure}
    \end{center}
    \begin{center}
     \begin{subfigure}{}
     \includegraphics[width=0.49\textwidth]{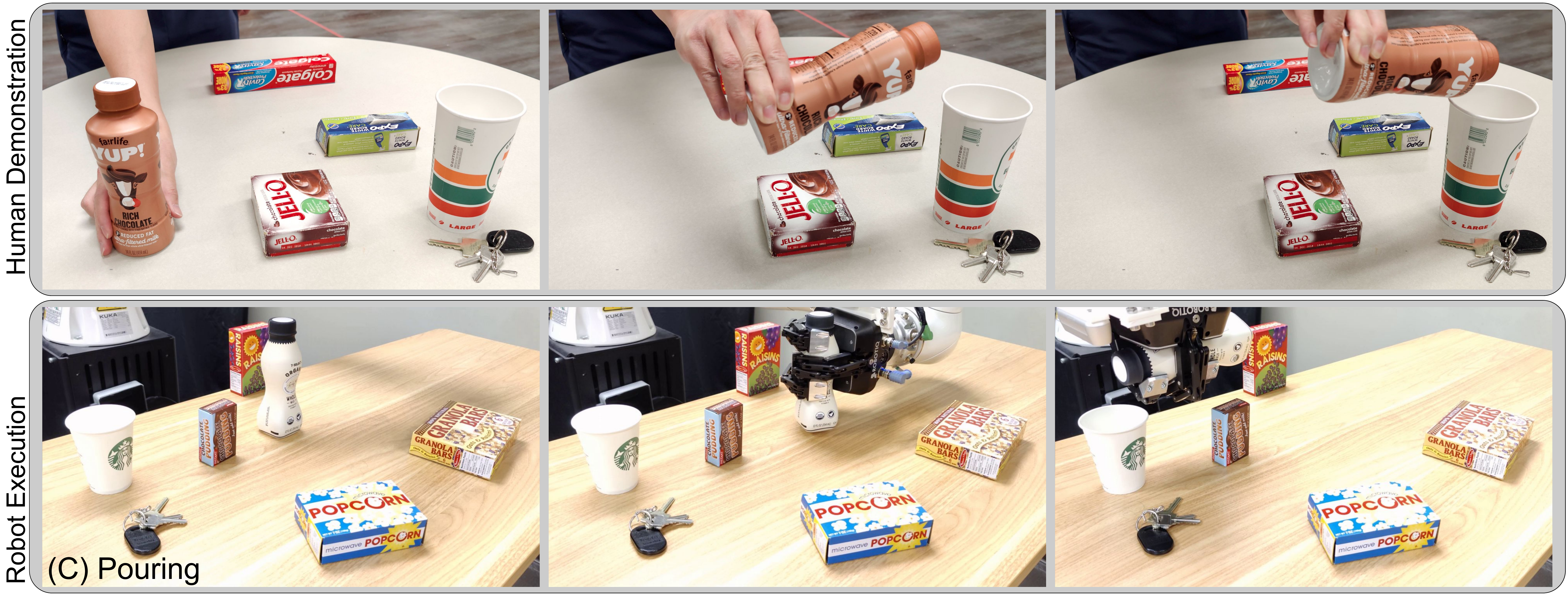}
     \end{subfigure}
     \hspace{-4mm}
     \begin{subfigure}{}
     \includegraphics[width=0.49\textwidth]{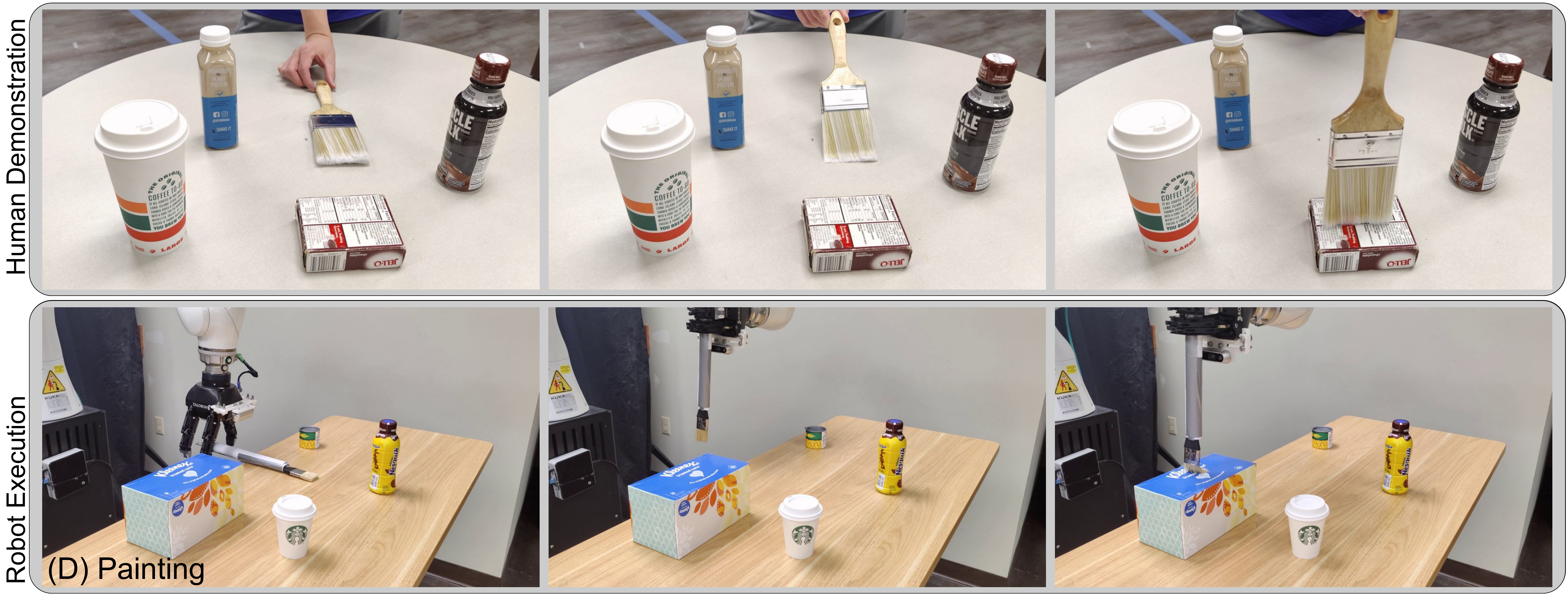}
     \end{subfigure}     
    \end{center}
     \caption{Examples of demonstration videos used for training (top rows) in each of the four manipulation tasks, and examples of the resulting robot executions on novel scenes (bottom rows).}
\label{fig:tasks}
\end{figure*}

{\bf Intermediate-level policy.} The pair of objects  $o_{\textrm{target}}$ and $o_{\textrm{tool}}$, received from high-level policy $\pi_h$, is provided as input to an intermediate-level policy $\pi_m$, which is 
a second neural network  with a structure identical to $F_{\theta}$, except for the last layer that returns a 6D keypose $K$ instead of a scalar. Returned keypose $K$ is the desired pose of the tool in the target object's coordinates system. For stacking, for example, desired keypose puts the tool object right on the surface of the target object. For painting, the keypose of a brush places the tip of the brush on the surface of another object that needs painting, and keeps the brush orthogonal to the target's surface. 
Keyposes are defined here as relative placements of objects with respect to each other, which are more consistent than the ones expressed in the camera coordinates system. Additionally, as only relative poses between objects are used here, one can easily use a different calibrated camera during robot execution (i.e. testing) without the need for an alignment with the camera pose used during the demonstrations.  

We use rotation matrices in input features $\Phi$ and a quaternion for the orientation in output $K$, because transformations with homogeneous matrices can be easily represented in linear operations inside neural networks, while the quaternion is a better output format as it requires less constraints. Similar to high-level policy $\pi_h$, $\pi_m$ is also class-agnostic, and keypose computation can be shared among similar-shaped objects, which facilitates the training. 

{\bf Low-level policy.} We use a standard motion planner from the {\it MoveIt} library as our low-level policy to move object $o_{\textrm{tool}}$ to its next desired keypose $K$ in the intrinsic frame of reference of object $o_{\textrm{target}}$, after receiving $o_{\textrm{tool}}$, $o_{\textrm{target}}$ and $K$ from the previous components. This is performed by computing a 
change $\Delta c_t$ to apply on the robot's configuration $c_t$, using an inverse kinematics model of the robot, pose of the target object in the robot's frame, desired pose $K$ of the tool, and the pose of the tool in the robot's frame. If $o_{\textrm{tool}}$ selected by the high-level policy happens to be the robot's hand, then the low-level policy systematically closes the hand once $o_{\textrm{tool}}$ is placed in $K$, which results in grasping the object. 

\subsection{Learning from Demonstrations}
During training, the system receives as inputs sequences of RGB-D images showing a human demonstrating a grasping-and-manipulation task on unknown objects. The system first automatically segments the images into individual point clouds, one per object, by removing the background. The objects include the human hand, for learning task-appropriate grasp poses. After segmenting the frames, an RGB-based tracker~\cite{Li_2018_CVPR} is applied to match segments across consecutive frames. The segment with the most significant motion is labeled as $o_{\textrm{tool}}$ unless that object is the hand of the demonstrator, in which case it is considered as the tool only if no object is being grasped.
The object closest to the tool object in the end of the demonstration is labeled as $o_{\textrm{target}}$, and the transformation $T_{o_\textrm{tool} \rightarrow o_\textrm{target}}$ in the end of the demonstration is considered as the final keypose $K$. This entire process if fully automated and does not require any human input other than performing the demonstration. 
Given these labels, we train the high-level policy network to maximize the likelihood of  $\pi_h(o_{\textrm{tool}}, o_{\textrm{target}})$. Intermediate-level policy $\pi_i$ is trained to minimize the mean squared error.

%% file: experiments.tex
\section{Experiments}

\begin{figure*}[h]
    \begin{center}
    \hspace{-8mm}
     \begin{subfigure}{}
     \includegraphics[width=0.24\textwidth]{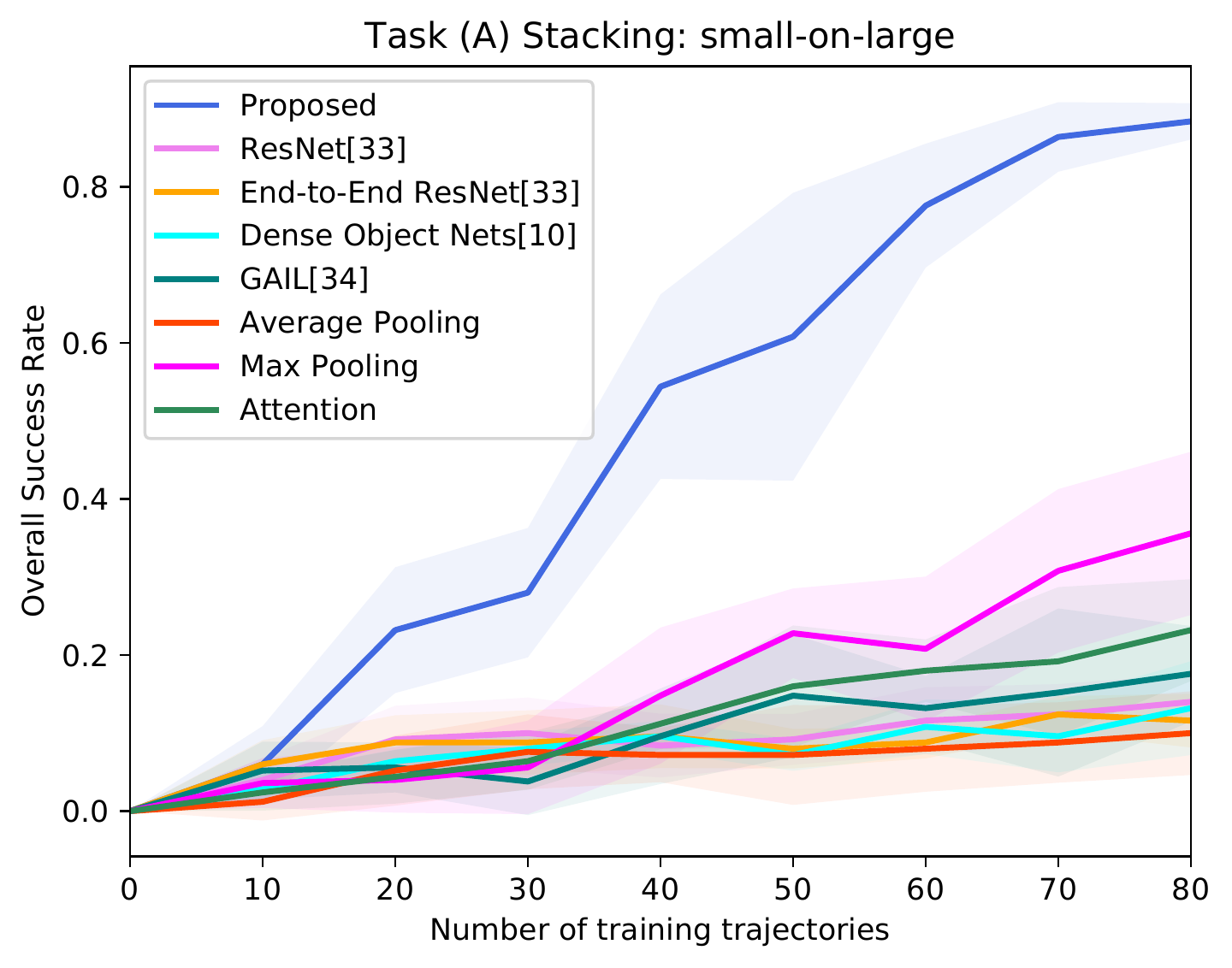}
     \end{subfigure}
     \hspace{-5mm}
     \begin{subfigure}{}
     \includegraphics[width=0.24\textwidth]{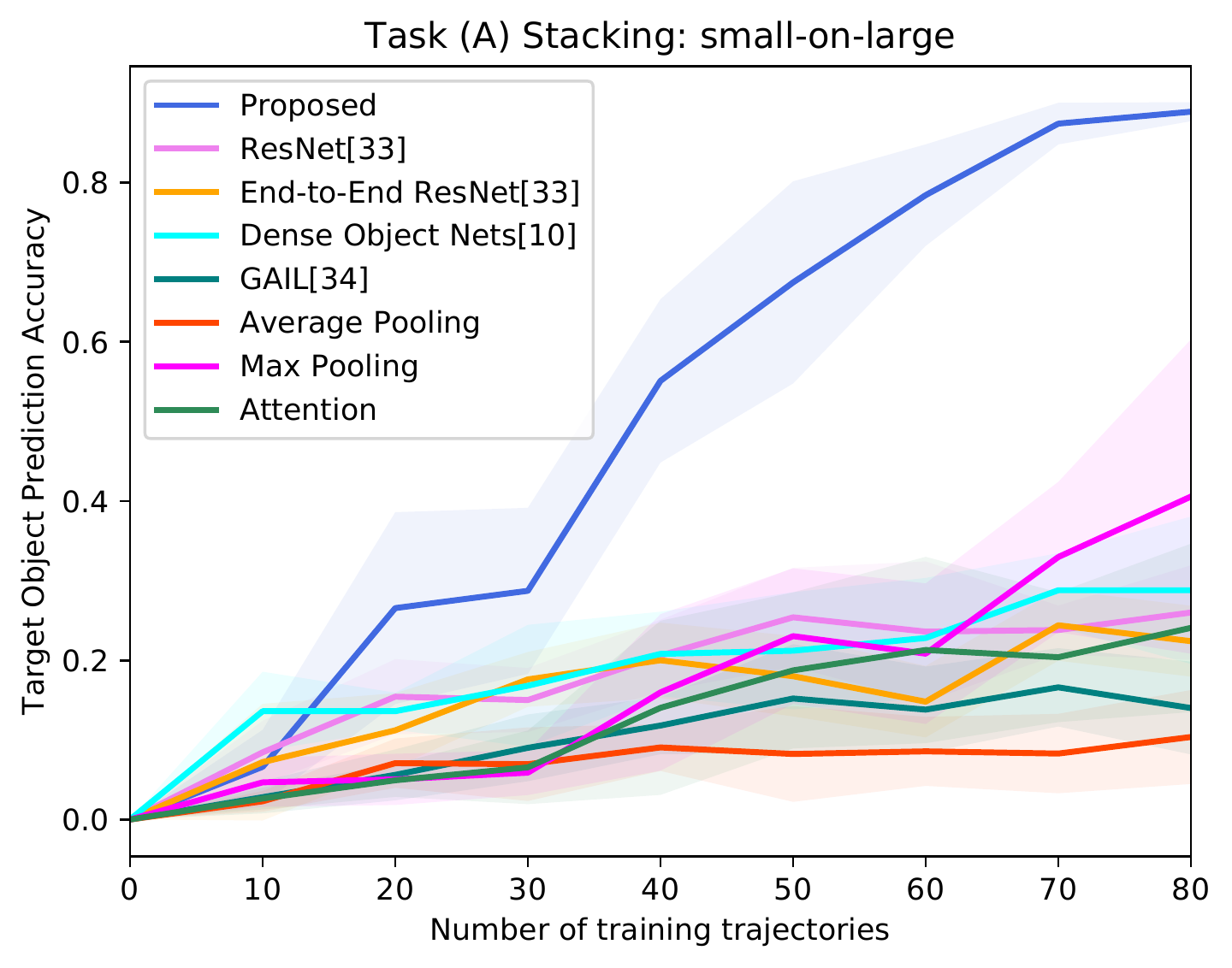}
     \end{subfigure}
     \hspace{-5mm}
     \begin{subfigure}{}
     \includegraphics[width=0.24\textwidth]{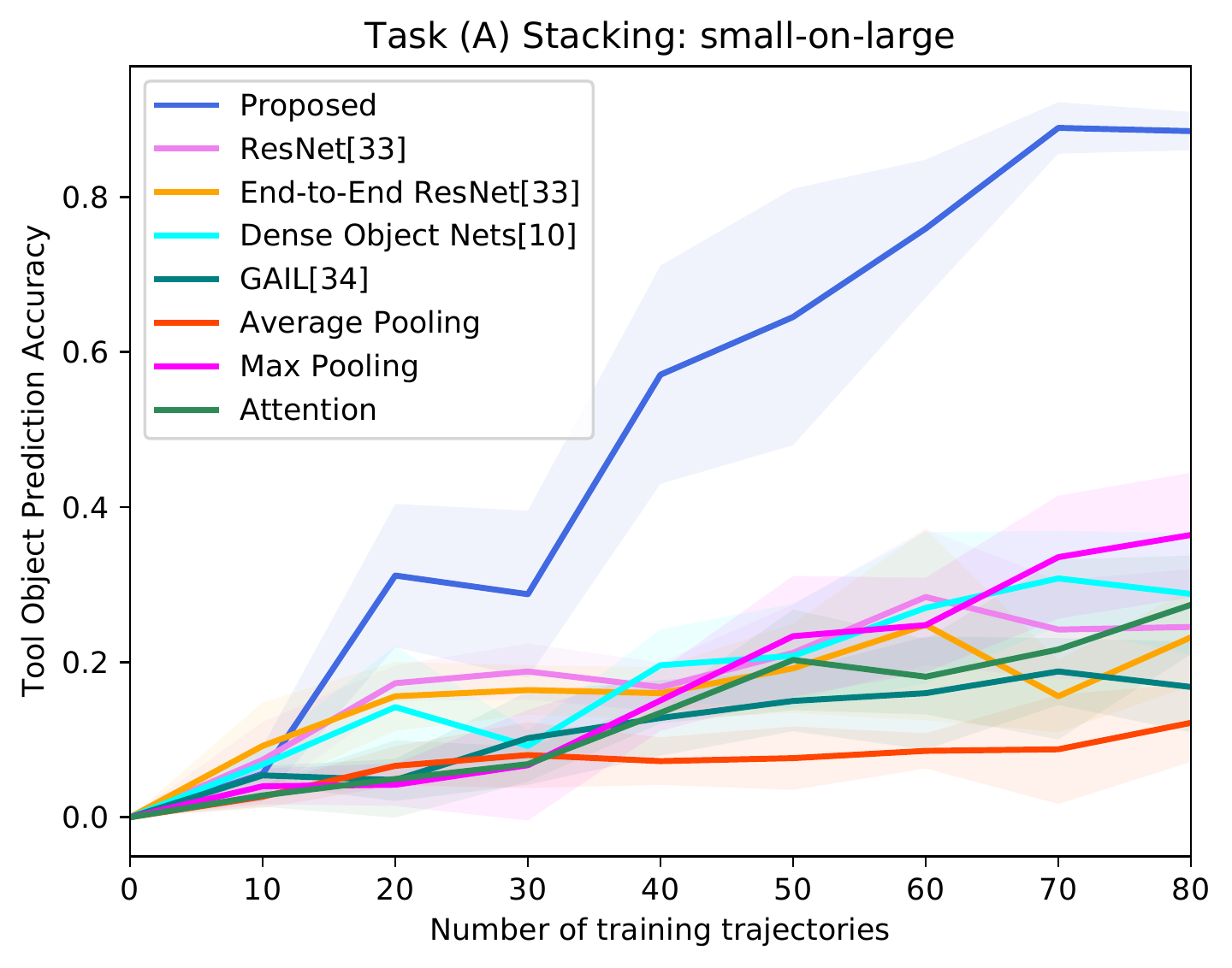}
     \end{subfigure}
     \hspace{-5mm}
     \begin{subfigure}{}
     \includegraphics[width=0.24\textwidth]{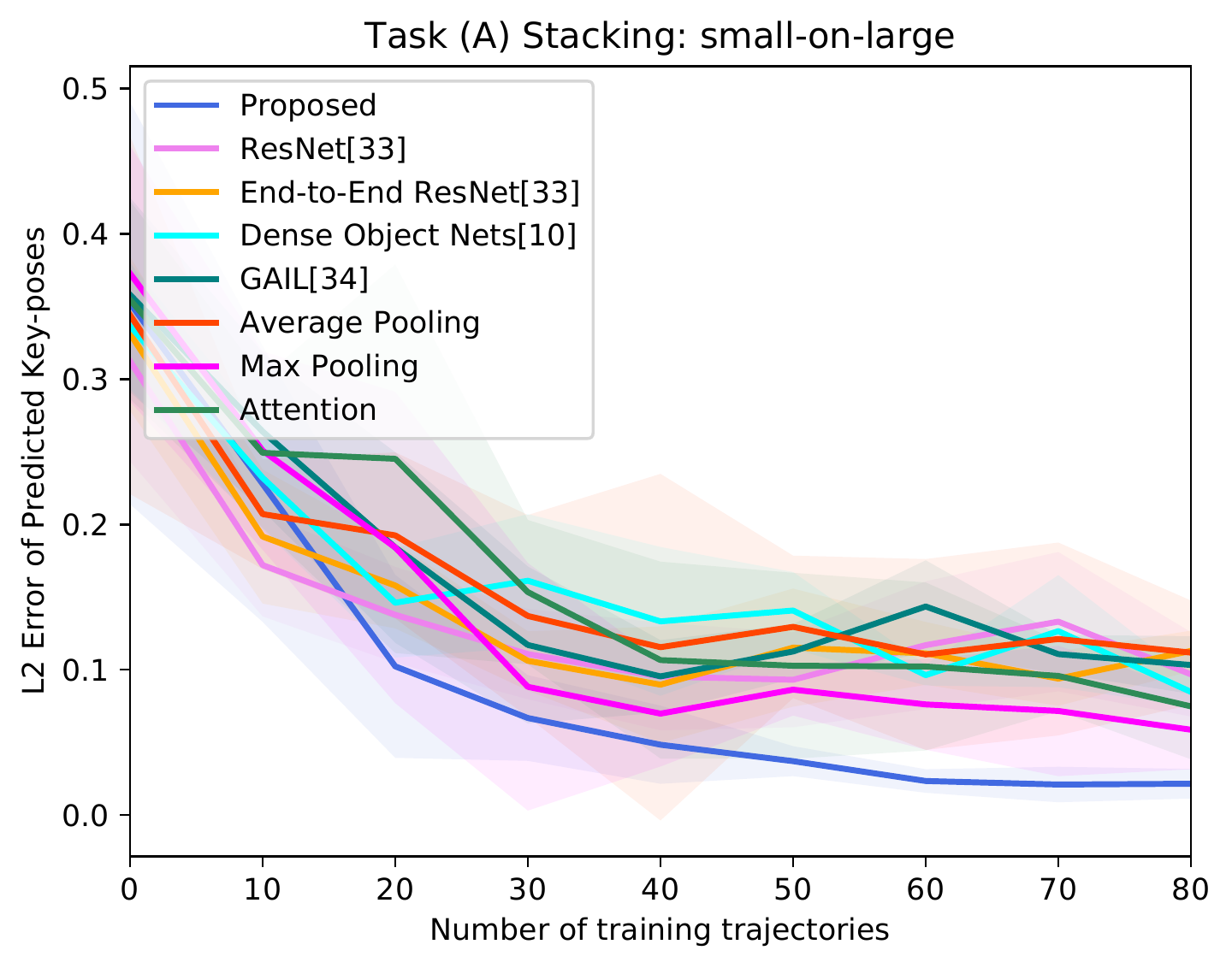}
     \end{subfigure}
     \end{center}
    \begin{center}
    \hspace{-8mm}
     \begin{subfigure}{}
     \includegraphics[width=0.24\textwidth]{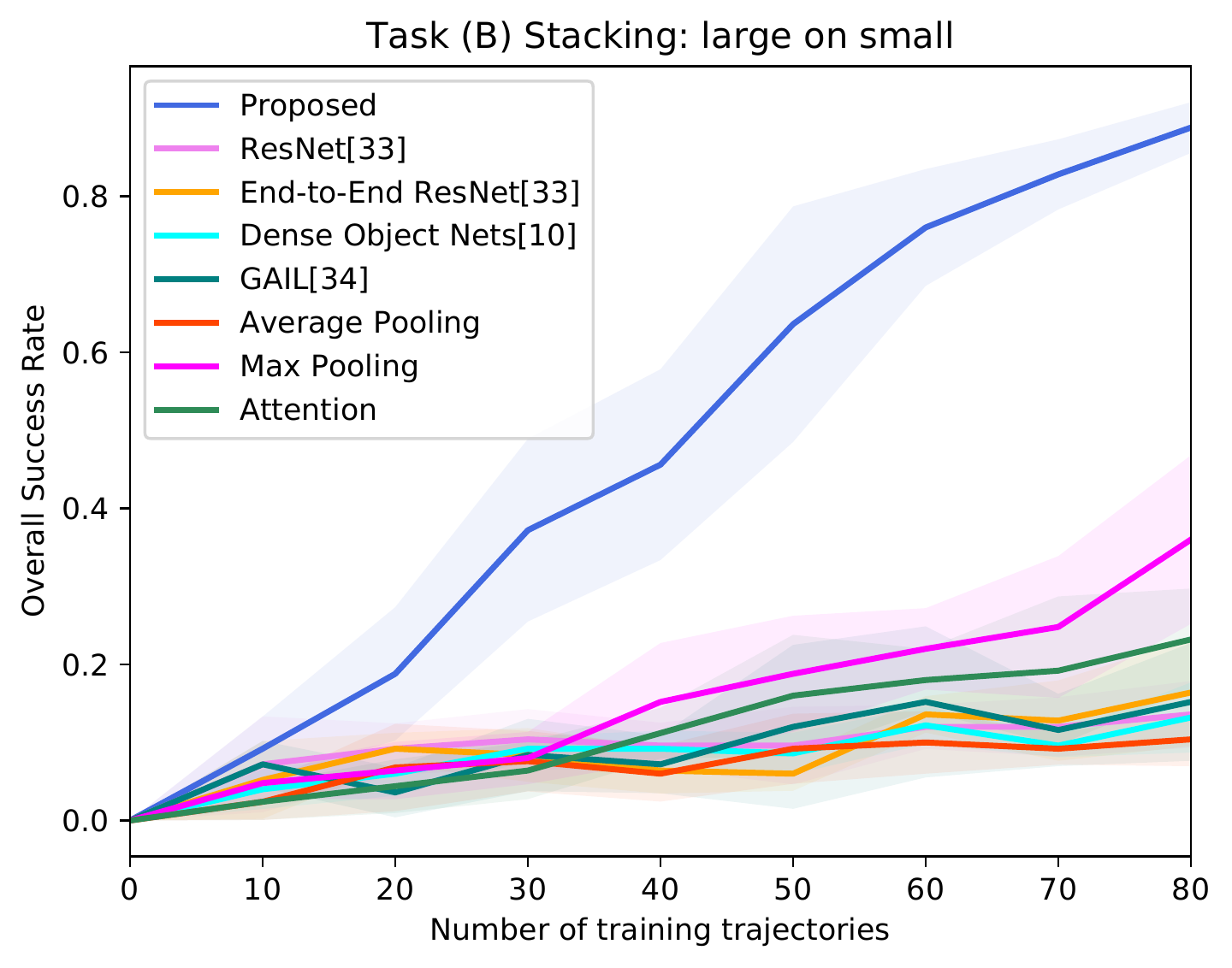}
     \end{subfigure}
     \hspace{-5mm}
     \begin{subfigure}{}
     \includegraphics[width=0.24\textwidth]{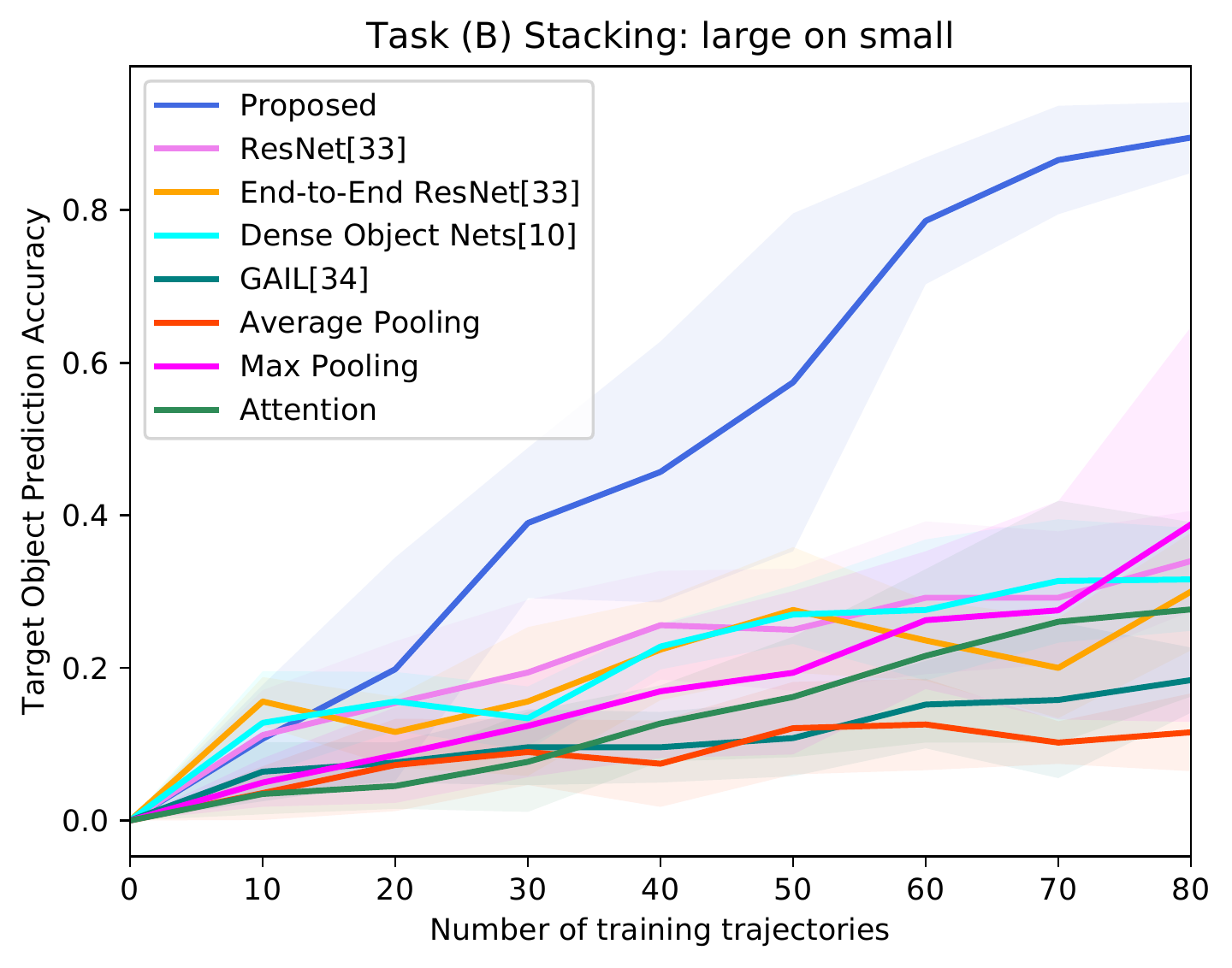}
     \end{subfigure}
     \hspace{-5mm}
     \begin{subfigure}{}
     \includegraphics[width=0.24\textwidth]{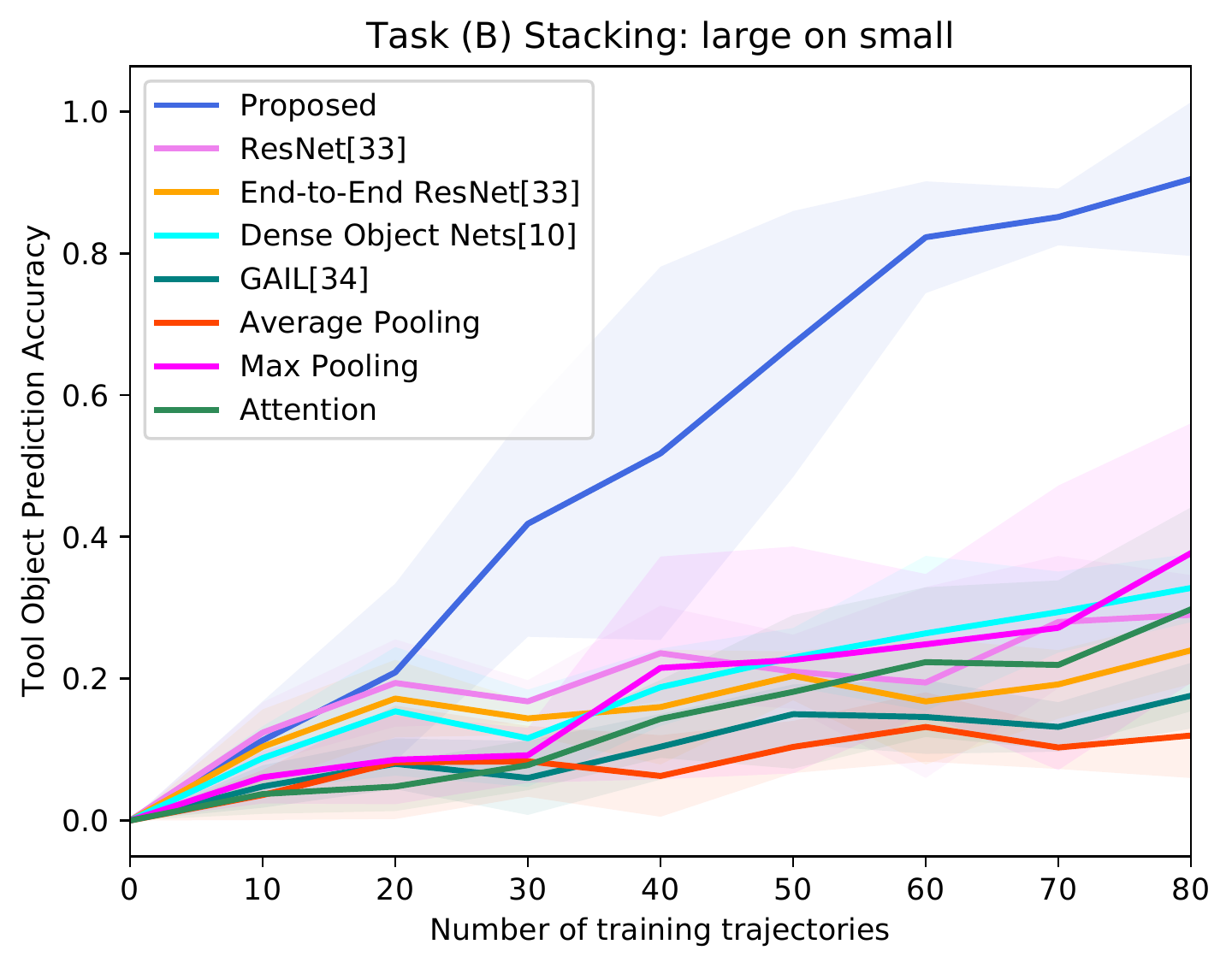}
     \end{subfigure}
     \hspace{-5mm}
     \begin{subfigure}{}
     \includegraphics[width=0.24\textwidth]{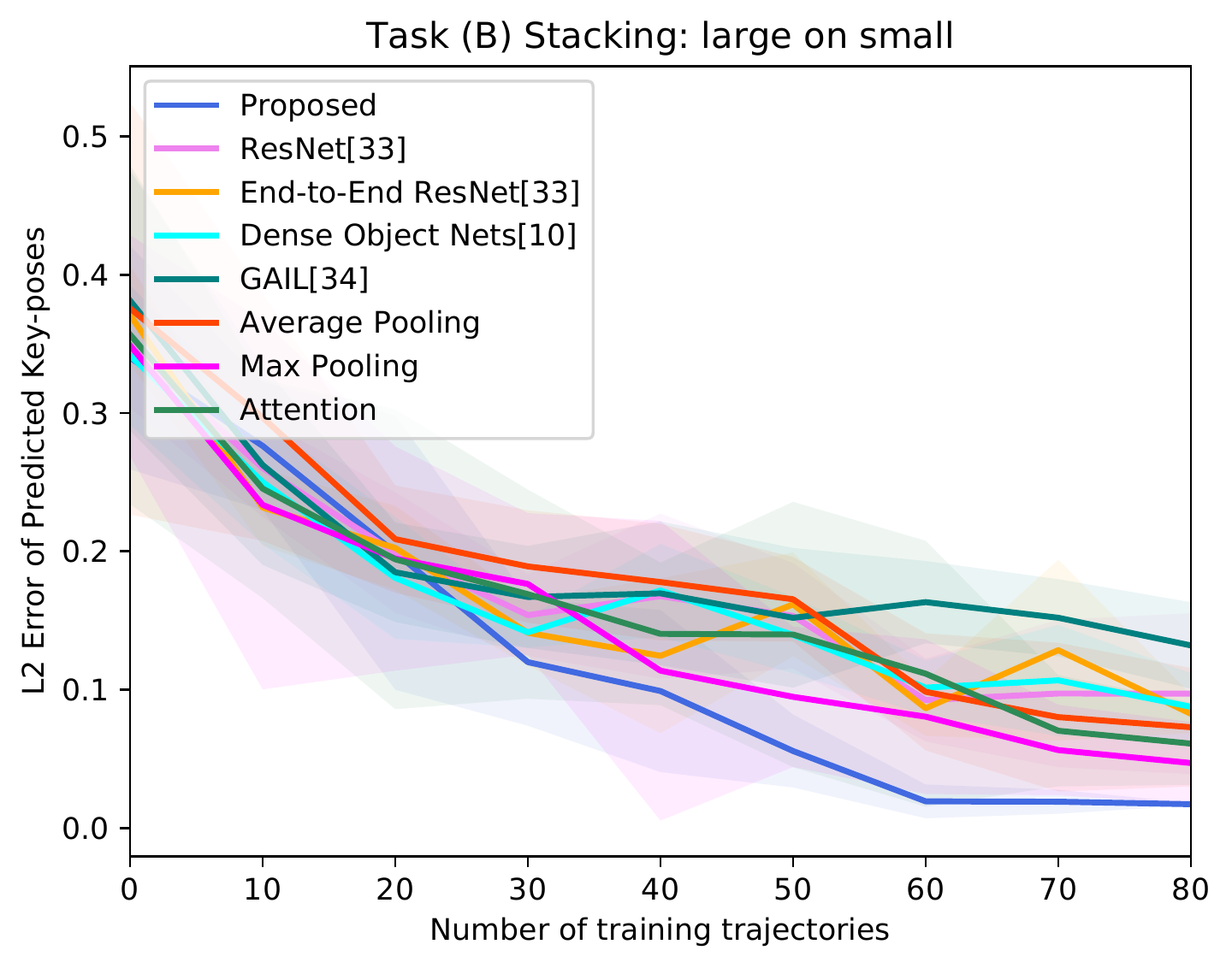}
     \end{subfigure}
     \end{center}
    \begin{center}
    \hspace{-8mm}
     \begin{subfigure}{}
     \includegraphics[width=0.24\textwidth]{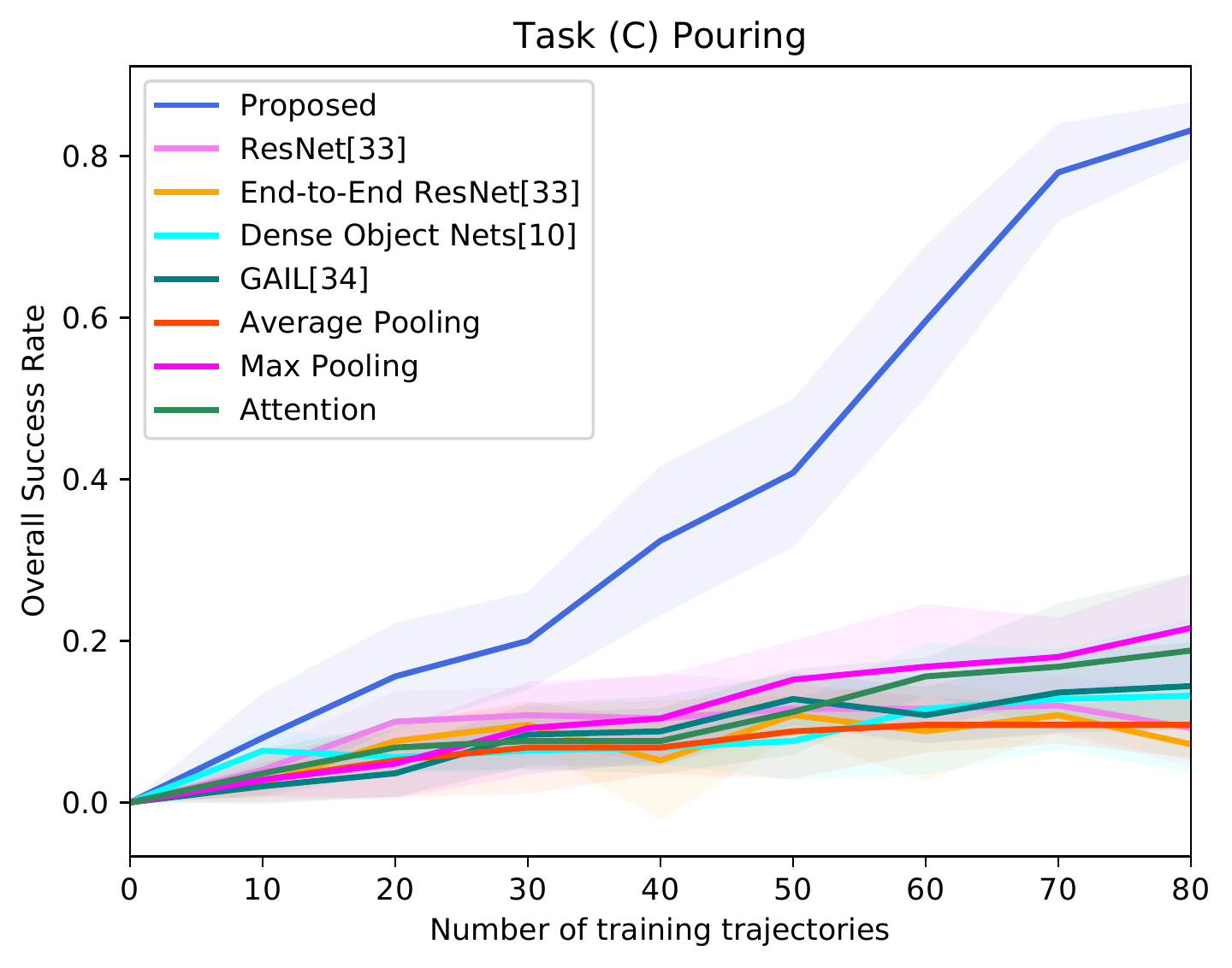}
     \end{subfigure}
     \hspace{-5mm}
     \begin{subfigure}{}
     \includegraphics[width=0.24\textwidth]{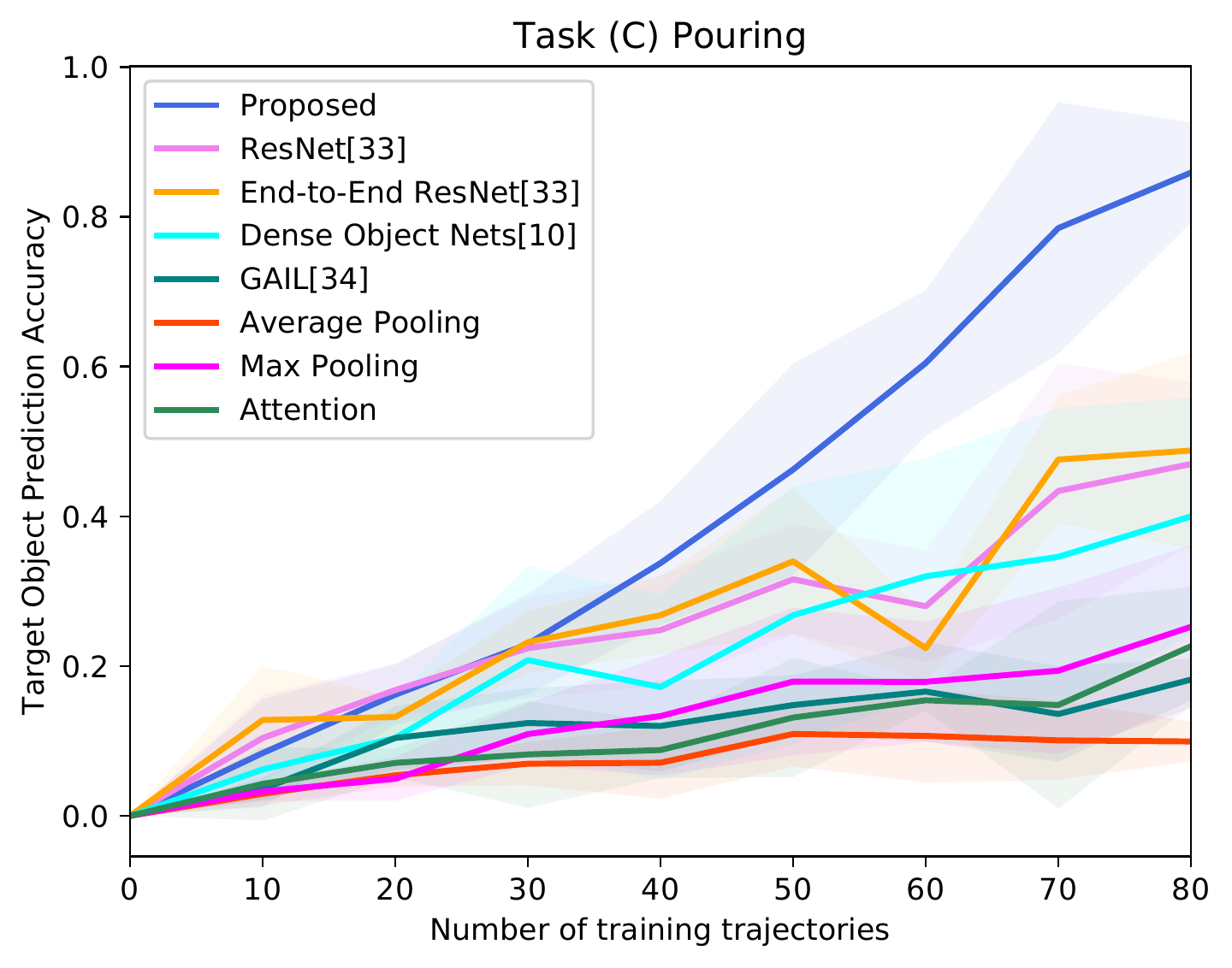}
     \end{subfigure}
     \hspace{-5mm}
     \begin{subfigure}{}
     \includegraphics[width=0.24\textwidth]{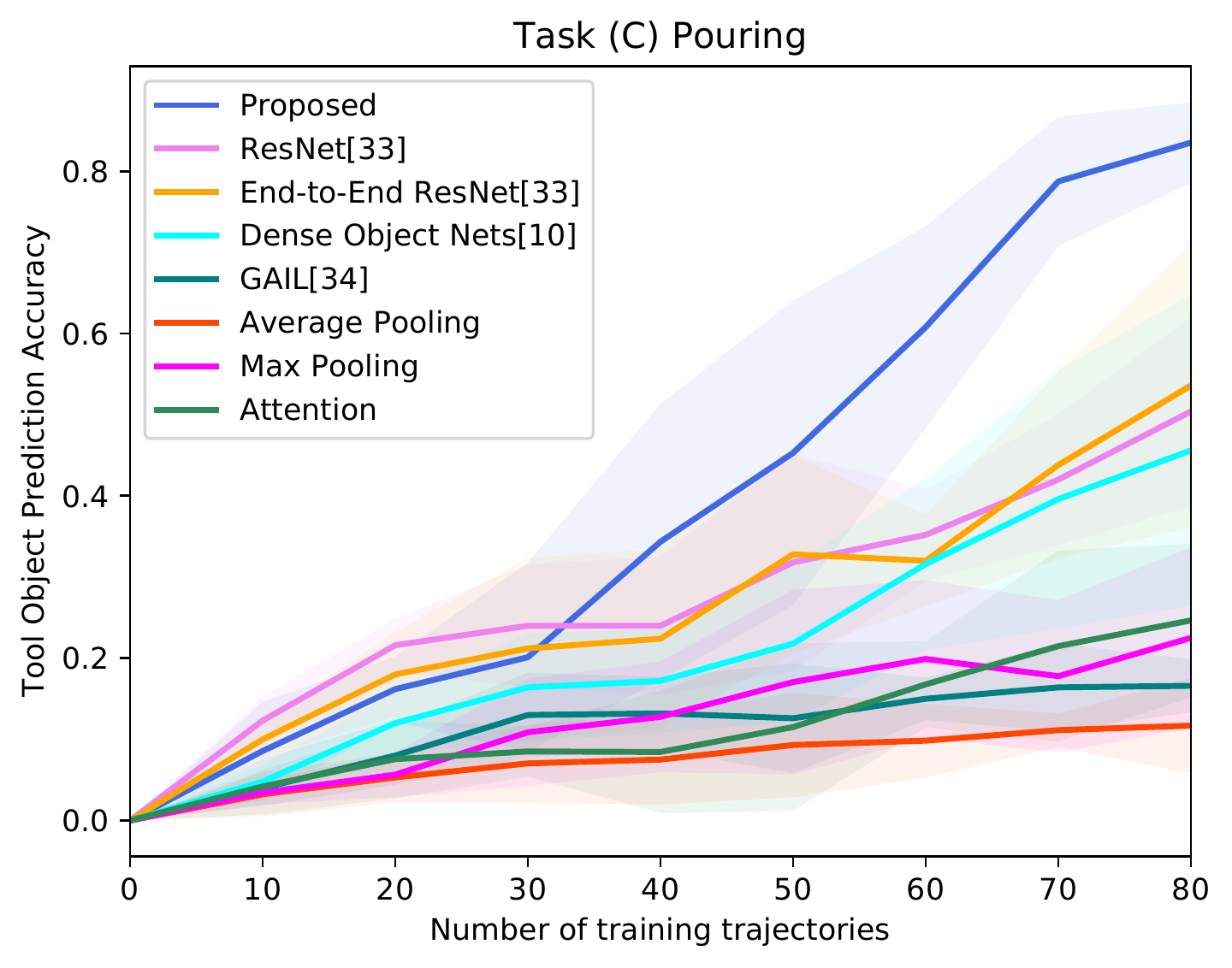}
     \end{subfigure}
     \hspace{-5mm}
     \begin{subfigure}{}
     \includegraphics[width=0.24\textwidth]{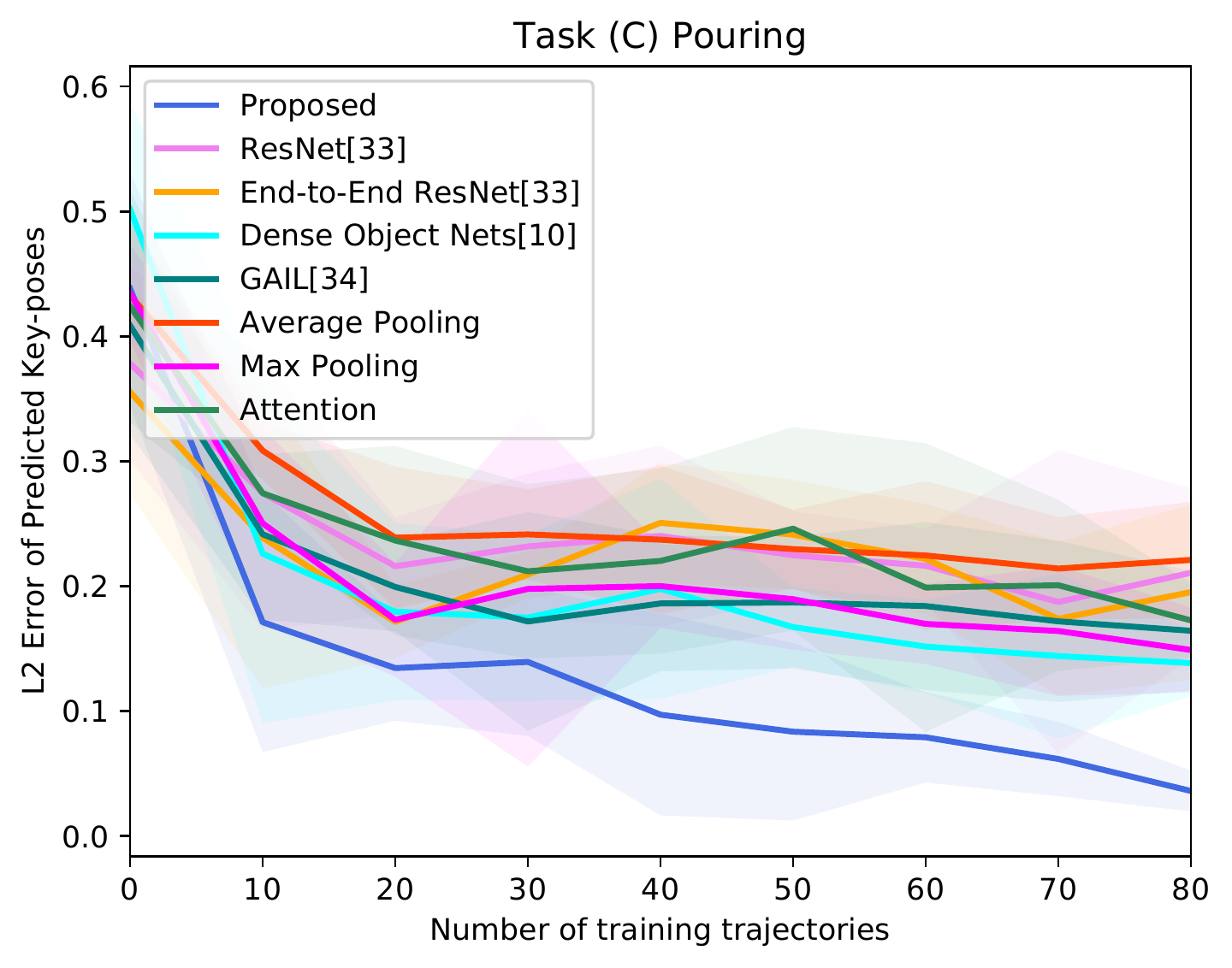}
     \end{subfigure}
     \end{center}
     
    \begin{center}
    \hspace{-8mm}
     \begin{subfigure}{}
     \includegraphics[width=0.24\textwidth]{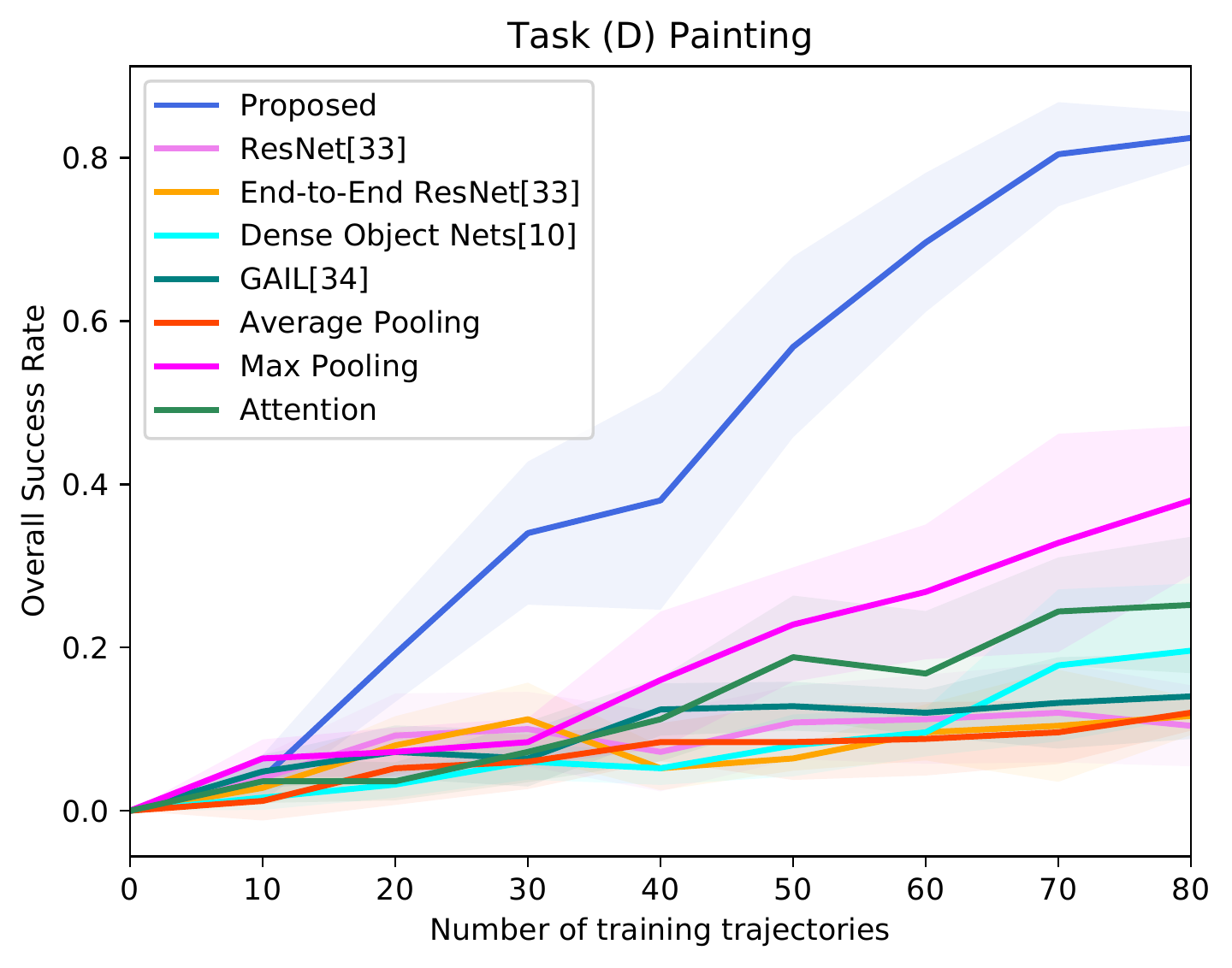}
     \end{subfigure}
     \hspace{-5mm}
     \begin{subfigure}{}
     \includegraphics[width=0.24\textwidth]{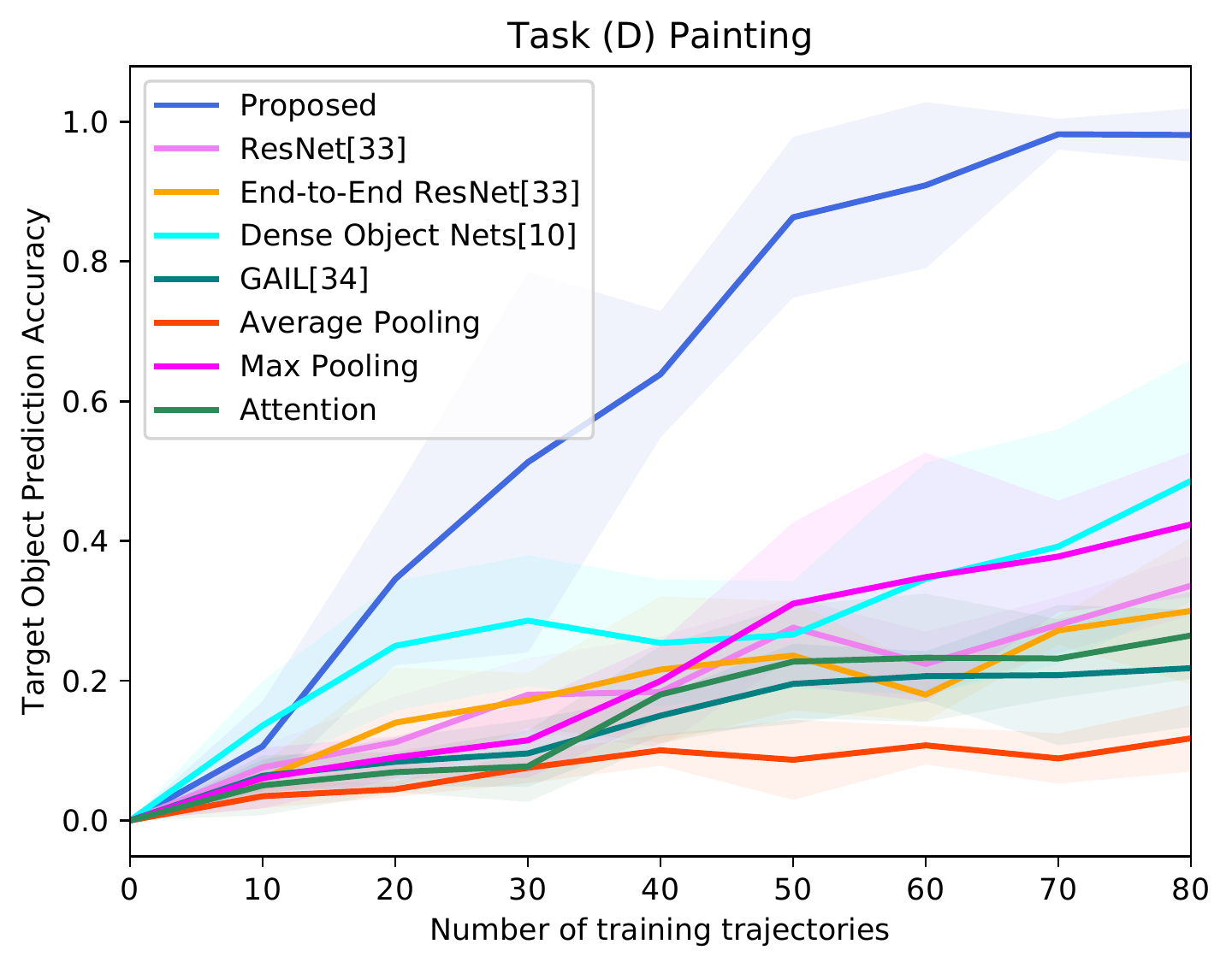}
     \end{subfigure}
     \hspace{-5mm}
     \begin{subfigure}{}
     \includegraphics[width=0.24\textwidth]{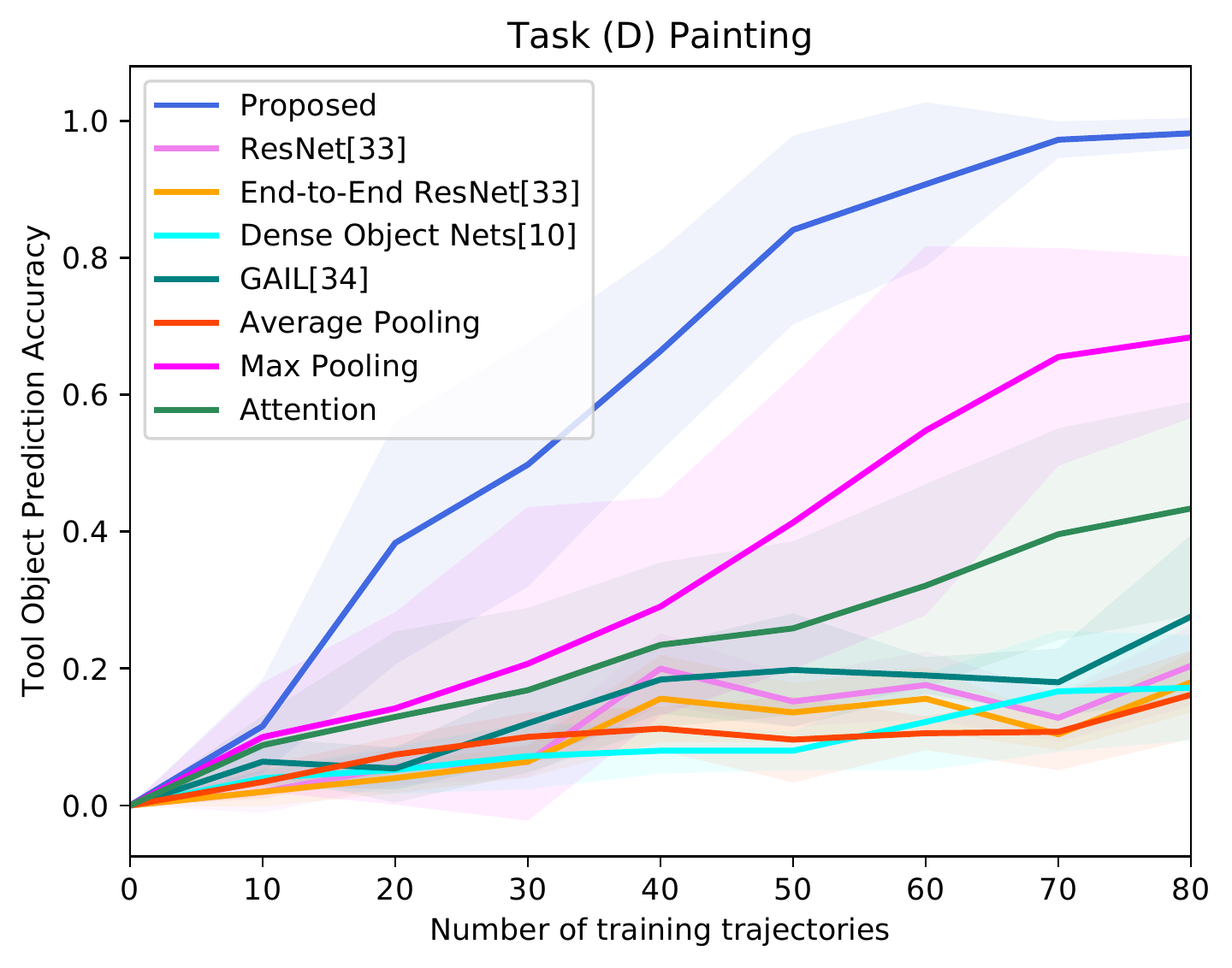}
     \end{subfigure}
     \hspace{-5mm}
     \begin{subfigure}{}
     \includegraphics[width=0.24\textwidth]{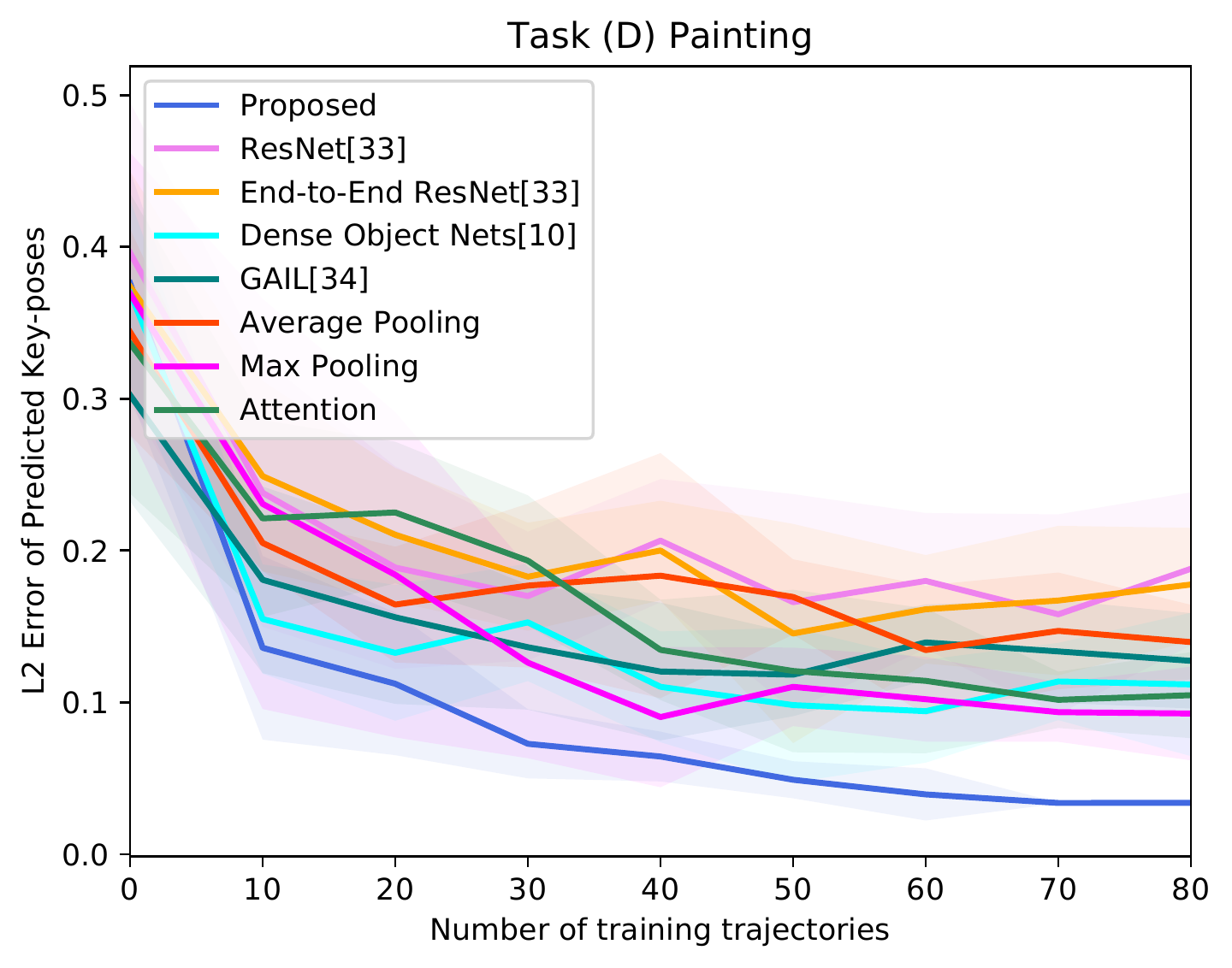}
     \end{subfigure}
     \end{center}
     
     \caption{Results of experiments on predicting tool-target pairs and keyposes in real novel static scenes, as a function of the number of videos used for training. The results are averaged over five independent trials.}
\label{fig:stat_all}
\end{figure*}

\begin{figure}
    \begin{center}
     \begin{subfigure}{}
     \includegraphics[width=0.49\textwidth]{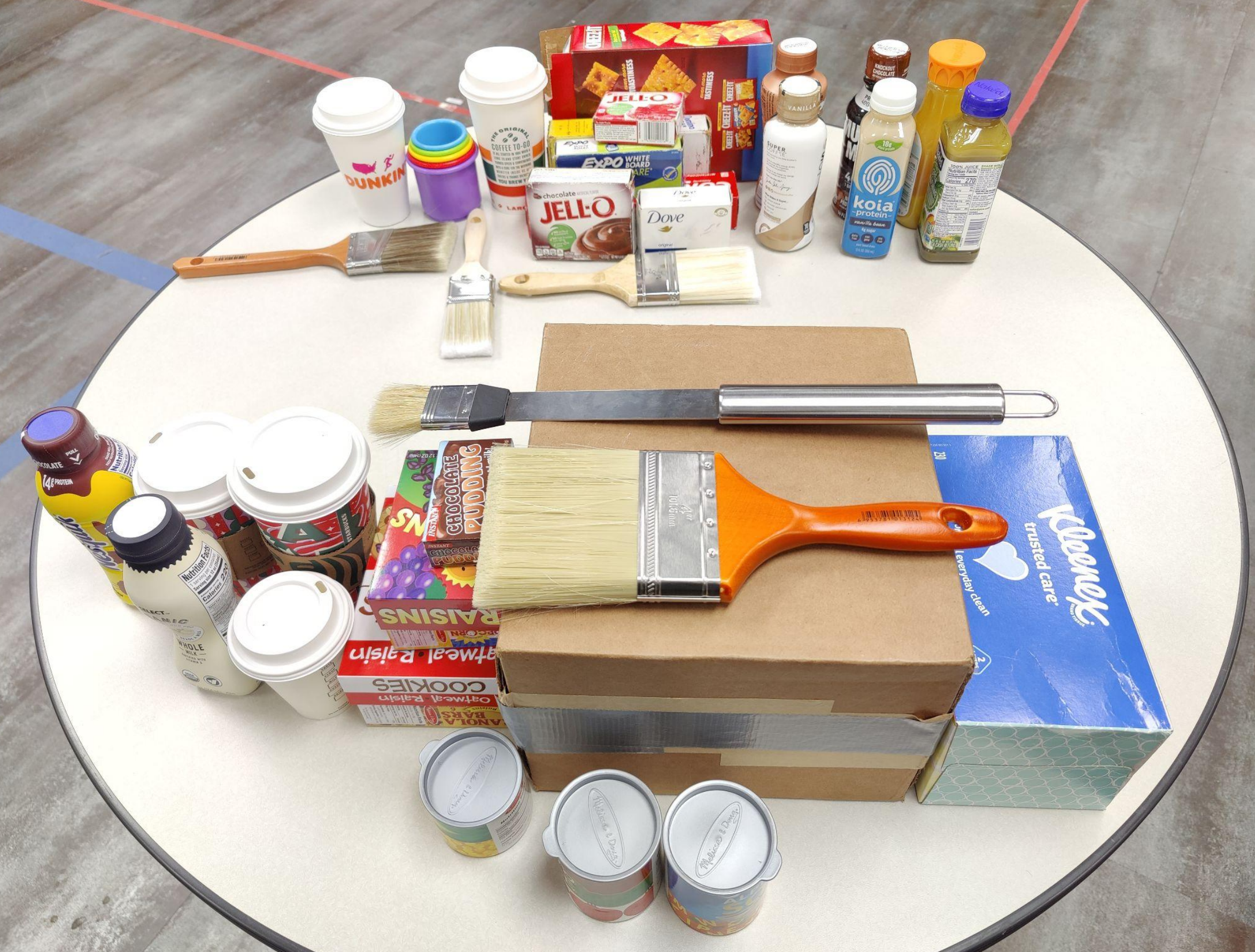}
     \end{subfigure}
    \end{center}
     \caption{Set of objects used in the experiments. Top row is the set of demonstration objects, bottom row is the test objects.}
\label{fig:obj_set}
\end{figure}

We evaluate the proposed method on four manipulation tasks, listed in Fig. \ref{fig:tasks}, and compare it with seven alternative architectures on the same tasks. In order to challenge the generalization capability of the trained systems in each task, we split a set of objects with various sizes and texture into a training set, used in demonstration videos, and a test set, used in robot executions, so that the objects in the test sets are never seen in the demonstrations, as shown in Fig.~\ref{fig:obj_set}. For each task, we record $80$ demonstration videos for training. The trained system is tested on $20$ different scenes per task.

 We compare the proposed architecture with methods that aggregate shape (from DGCNN) and appearance (from Fast-RCNN) descriptors of all the objects that are present in the scene into one large vector, and use that as an input to a neural network, which  returns a predicted target object, tool object and keypose. This is a more straightforward design that is different from our architecture that takes pairs of candidate tool-target as inputs, and decomposes the policy into a high-level and an intermediate-level ones. All compared methods share the same low-level policy.


The three compared architectures aggregate input object descriptors by: (1) {\bf average pooling}, (2) {\bf max pooling}, and (3) {\bf attention}, respectively. Without the flexibility in predicting the target and the tool object in  scenes made of variable numbers of objects, the baselines use instead two classification output layers for predicting target and tool object separately. Returned tool and target objects are indicated by their indices in the input. 
Since the maximum number of objects per scene in our experiments is  $10$, the classification output layers of the baseline models have $10$ units. This is a limitation compared to our proposed architecture, which is class-agnostic and can handle scenes with any number of objects. 
 Comparisons in these experiments should answer two questions: (1) how much improvement the DGCNN feature can bring to category-level manipulation, (2) how significant is the proposed architecture. To answer the first one, the proposed method is compared with ResNet~\cite{he2016deep} and Dense Object Nets~\cite{DBLP:conf/corl/FlorenceMT18}. Besides feeding fixed pre-trained features, we also evaluate an end-to-end training. These baselines share the same architecture as the proposed one except the input features.

To answer the second question, we compare our architecture with a more straightforward design: these compared methods aggregate shape (from DGCNN) and appearance (from Fast-RCNN) descriptors of all the objects that are present in the scene into one large vector, and use that as an input to neural networks, which return a predicted target object, tool object and keypose. We evaluate three compared mechanisms in aggregating input object descriptors: {\bf average pooling}, {\bf max pooling}, and {\bf attention}. Without the flexibility in predicting the target and the tool object in  scenes made of variable numbers of objects, the baselines use instead two classification output layers for predicting target and tool object separately. Returned tool and target objects are indicated by their indices in the input. All compared methods share the same low-level policy. Since the maximum number of objects per scene in our experiments is  $10$, the classification output layers of the baseline models have $10$ units. This is a limitation compared to our proposed architecture, which is class-agnostic and can handle scenes with any number of objects. Additionally, to see how the proposed decomposition of the policy influences the training, we evaluate GAIL~\cite{ho2016generative} with an architecture similar with the max pooling described above, except that it predicts change of tool object poses in each time step rather than the keypose. Without the intermediate-level policy, this architecture need to predict tool object, target object and low-level motion simultaneously. GAIL is an inverse imitation learning algorithm which have been proven successful in many applications. But even with this training algorithm, the lack of the decomposition of its policy still blocks this baseline from solving tasks shown later.

All compared methods use two consecutive fully connected layers with $256$ and $128$ units to encode object descriptors. In all models, the size of the hidden fully connected layer is $128$, and another hidden layer with $64$ units is followed before the quaternion output. ReLU is applied after every fully connected layer except for the output layers. We train all models with Adam optimizer, a training batch contains $10$ scenes, the learning rate is $1e-4$, and the number of epochs is $100$.

In all four tasks, we evaluate the methods based on their: (1) accuracy in predicting  the ground-truth target-tool pairs in each scene, (2) accuracy in predicting  keyposes of the tool in each scene, (3) overall task success rate, which combines together the two previous criteria. 
The keypose prediction accuracy is measured by the $L_2$ error. 
Additionally, we deploy the trained models on a {\it Kuka iiwa14} robot equipped with a {\it RealSense 415} RGB-D camera (Fig.~\ref{fig:system_overview} and Fig.~\ref{fig:tasks}) and report the task success rates in table \ref{tab:real_robot_stat}.


\begin{table}[h]
    \centering
    \begin{tabular}{c|c|c|c|c}
         Success / Trial & Task (A)  & Task (B) & Task (C) & Task (D) \\
         \hline
         Proposed & $5/5$ & $5/5$ & $4/5$ & $4/5$\\
         \hline
         Average Pooling & $0/5$ & $0/5$ & $0/5$ & $0/5$\\
         \hline
         Max Pooling & $2/5$ & $1/5$ & $1/5$ & $0/5$ \\
         \hline
         Attention & $1/5$ & $1/5$ & $1/5$ & $1/5$ \\
            \hline
         ResNet~\cite{he2016deep} & $0/5$  & $1/5$ & $1/5$  & $0/5$ \\
            \hline
         End-to-End ResNet~\cite{he2016deep} & $0/5$  & $1/5$ & $0/5$  & $0/5$ \\
            \hline
         GAIL~\cite{ho2016generative} & $0/5$  & $0/5$ & $0/5$  &  $0/5$\\
            \hline
         Dense Object Nets~\cite{DBLP:conf/corl/FlorenceMT18} & $1/5$  & $0/5$ & $0/5$  &  $0/5$ \\
    \end{tabular}
    \caption{Real robot evaluation results}
    \label{tab:real_robot_stat}
\end{table}

\noindent {\bf Small-on-large box stacking (A), and large-on-small box stacking (B).} The robot is expected to stack the smallest box on top of the largest one in task (A) and do the opposite in task (B). The desired ideal keypose for placement is on top of the center of the target object, with its height given as the sum of the target object's height and half of the tool object's height. A predicted keypose is considered as accurate as long as its pose is within $3cm$ from the ideal center. The result of these two tasks are shown in top two rows of Fig. \ref{fig:stat_all}. We can see from these plots that the proposed model achieves better performance than compared models. While the proposed model gets lower error in keypose prediction, its advantage is even more significant in the target and tool object prediction. We hypothesize that the keypose prediction in these two tasks is easier than tool-target prediction because the keypose centers are always on top of the target objects' centers in the training videos. Furthermore, ResNet and Dense Object Nets baselines, which feed compared features to the proposed architecture, also reach higher accuracy on target and tool object prediction than baselines with other aggregation mechanisms. So results on these tasks support the benefit of proposed architecture, as an answer to the significance of the proposed architecture. It can be explained by the fact that the order of objects is arbitrary in the different scenes. Therefore, the baseline models without proposed architecture will see many different indices for the same object in different scenes during training. As a result, training iterations can cancel each other until these models manage to find out the correspondence between each input object and each output. In contrast, the proposed model can avoid this problem because the same object in difference scenes or frames should get similar outputs as they have similar descriptors. Among the compared aggregation mechanisms, the average pooling model seems to fail to learn. It may fail to focus on target and tool objects because it always considers all the objects in the scene. Although they may need much more data than our model, the max pooling and attention models show improving tendencies. Max pooling seems to be better than the attention model, its relative advantage may come from the fact that (1) it has less parameters as it does not compute attention weights, and (2) these tasks require only two objects rather than a complex combination of information from multiple objects. After verifying the significance of the proposed architecture, we can see the benefit of DGCNN features through the difference between the proposed method with ResNet and Dense Object Nets baselines. Without direct shape information from point clouds, these baselines have higher keypose prediction errors than other evaluated methods. At the same time, they are more likely to fail distinguishing between large and small box as object appearances can be misleading in sizes due to their distances to the camera. Additionally, the end-to-end training does not seem to help here.

\noindent {\bf (C) Pouring.} The robot is expected to pick up a bottle from the side,  move it close to a cup, and rotate it to point to the cup. The ideal desired keypose has a height defined as the sum of the height of the cup and the radius of the bottle, and a successful pose should be higher than the cup and less than $5cm$ + desired height. In addition to the height criteria,  a successful pose should also satisfy: (1) the pose has a main axis in its PCA pose pointing toward the cup, the direction of this axis goes from the bottle's center to the bottle's opening (2) the point cloud of the bottle's neck  transformed by the predicted key pose lies within the range of the cup, the size of the cup is computed by its point cloud. Performances of different models are shown in the third row of Fig. \ref{fig:stat_all}. We can also see the clear advantage of the proposed model over other models, and the advantage in keypose predictions is more significant than in previous tasks, which should come from the fact that the keypose in this task is more complex: In this task, the ground-truth keypose depends on the initial relative pose of the bottle with respect to the cup because the demonstrator pours from left if the bottle is on the left of the cup initially, and pours from the right otherwise. Our intermediate-policy can extract this information from the input relative poses, while this information cannot be retrieved directly from the input with other models. Experiments also show gaps in target and tool object prediction between proposed method and compared input features, ResNet and Dense Object Nets, are smaller in this task. This result can be attributed to the appearance saliency of bottles and cups over other objects. But this still cannot reduce their disadvantages in keypose prediction as 3D shape information is missing in these features.

\noindent {\bf (D) Painting.}  The robot is expected to lift and rotate a paint-brush, and move it to the top surface of a box. The desired keypose requires the brush's tip to perpendicularly touch the top surface of the box. As the brush's tip is soft, we allow a tolerance in height of $3cm$.  Performances of different models are listed in the last row of Fig. \ref{fig:stat_all}, which again shows a clear advantage of the proposed method. But we can see that in this case, max pooling and attention achieve good accuracy in target object prediction. We hypothesize that the reason is that brushes have much different shapes than other objects present in the scene, and hence, the tool object is easier to be distinguished when DGCNN features are provided. Furthermore, low accuracy for target and tool object prediction from compared features, ResNet and Dense Object Nets, again suggests that appearance information alone can be insufficient for these manipulation tasks.

%% file: conclusion.tex
\section{Conclusion}

We presented a novel robot imitation learning framework for performing manipulation tasks in scenes that contain multiple unknown objects. The proposed model takes features from images and point clouds as input, and predicts a pair of target and tool objects, and a desired keypose for placing the tool relative to the target. The proposed system does not require any predefined class-specific priors, and can generalize to new objects within the same category. Extensive experiments using real demonstration videos and a real robot show that the proposed model outperforms alternative models. Supplementary material, including video, can be found at   \href{ https://tinyurl.com/2swbt7a3 }{\texttt{\textcolor{blue}{ https://tinyurl.com/2swbt7a3 }}}.

\section*{Acknowledgement}
This work was supported by NSF awards IIS 1734492 and IIS 1846043.